\def\cleanrevisions{1}

\documentclass{article}

\usepackage{iclr2027_conference,times}

\usepackage[utf8]{inputenc}
\usepackage[T1]{fontenc}
\usepackage{hyperref}
\usepackage{url}
\usepackage{booktabs}
\usepackage{amsfonts}
\usepackage{nicefrac}
\usepackage{microtype}
\usepackage{xcolor}

\usepackage{algorithm}
\usepackage{algpseudocode}

\usepackage{graphicx}
\usepackage{fontawesome5}
\DeclareRobustCommand{\githubcodelink}{\href{https://gabrieleserussi.github.io/vedje/}{\textcolor{black}{\raisebox{-0.1ex}{\faGithub}}}}
\usepackage{subcaption}
\usepackage{caption}
\usepackage{bbm}
\usepackage{amsmath}
\usepackage{amssymb}
\usepackage{mathtools}
\usepackage{amsthm}
\usepackage{array}
\usepackage{adjustbox}
\usepackage{multirow}
\usepackage{colortbl}
\usepackage{enumitem}
\usepackage{needspace}
\usepackage{placeins}

\definecolor{ourmethodrow}{RGB}{255, 230, 230}
\newcommand{\oursrow}{\rowcolor{ourmethodrow}}

\DeclareRobustCommand{\methodcite}[1]{\space{\normalsize\citep{#1}}}
\DeclareRobustCommand{\methodcell}[2]{#1~\citep{#2}}
\newcolumntype{O}{>{\columncolor{ourmethodrow}}c}

\usepackage{float}
\usepackage{tabularx}
\usepackage{xspace}
\usepackage{tcolorbox}
\usepackage{pifont}

\usepackage{tikz}
\usetikzlibrary{arrows.meta,positioning,decorations.pathreplacing}

\newcommand{\cmark}{\checkmark}%
\newcommand{\xmark}{\ding{55}}%
\newcommand{\deltaloss}{\ensuremath{\mathcal{L}_{\mathrm{delta}}}\xspace}

\newif\ifshowrevisions
\ifdefined\cleanrevisions
  \showrevisionsfalse
\else
  \showrevisionstrue
\fi
\definecolor{revisionblue}{HTML}{B45309}
\DeclareRobustCommand{\rev}[1]{%
  {\ifshowrevisions\color{revisionblue}\else\color{black}\fi#1}
}
\newenvironment{revisionblock}
  {\begingroup\ifshowrevisions\color{revisionblue}\else\color{black}\fi}
  {\endgroup}

\definecolor{revisionorange}{HTML}{B45309}
\colorlet{editcolor}{revisionorange}
\DeclareRobustCommand{\edit}[1]{%
  {\ifshowrevisions\color{revisionorange}\else\color{black}\fi#1}
}
\newenvironment{editblock}
  {\begingroup\ifshowrevisions\color{revisionorange}\else\color{black}\fi}
  {\endgroup}
\usepackage{etoolbox}
\expandafter\def\csname revisedbib@li2023unmasked\endcsname{1}
\expandafter\def\csname revisedbib@bolya2025perceptionencoder\endcsname{1}
\AtBeginEnvironment{thebibliography}{%
  \let\originalbibitem\bibitem
  \RenewDocumentCommand{\bibitem}{o m}{%
    \par\ifshowrevisions
      \ifcsname revisedbib@#2\endcsname\color{revisionorange}\else\color{black}\fi
    \else\color{black}\fi
    \IfNoValueTF{#1}{\originalbibitem{#2}}{\originalbibitem[#1]{#2}}%
  }%
}
\AtEndEnvironment{thebibliography}{\color{black}}

\usepackage{wrapfig}

\usepackage[capitalize,noabbrev]{cleveref}

\definecolor{commgreen}{RGB}{34,139,34}

\algrenewcommand\algorithmiccomment[1]{\hfill // #1}
\algrenewcommand\algorithmicrequire{\textbf{Input:}}
\algrenewcommand\algorithmicensure{\textbf{Output:}}

\definecolor{archcolor}{HTML}{ff2800}
\newcommand\architecture{VEDJE}
\DeclareRobustCommand{\vedjevariant}[2]{%
  \ensuremath{%
    \text{\architecture{}}_{\mathrm{#1}}%
    \ifnum\pdfstrcmp{#2}{64}=0 \relax
    \else
      \text{-}#2%
    \fi
  }%
}
\DeclareRobustCommand{\vedjevariantbf}[2]{%
  \ensuremath{%
    \text{\textbf{\architecture{}}}_{\mathbf{#1}}%
    \ifnum\pdfstrcmp{#2}{64}=0 \relax
    \else
      \text{\textbf{-#2}}%
    \fi
  }%
}

\title{VEDJE: Video-Efficient Discriminative Joint\\Encoder for Scalable Video-Text Retrieval}

\author{%
Shahaf Wagner\textsuperscript{1,*}\enspace
Gabriele Serussi\textsuperscript{1,*}\enspace
Dan Ben Ami\textsuperscript{1}\enspace
Tomer Galanti\textsuperscript{2}\enspace
Chaim Baskin\textsuperscript{1}\\
{\normalfont\small\textsuperscript{1}INSIGHT Lab, Ben-Gurion University of the Negev, Israel}\\
{\normalfont\small\textsuperscript{2}Texas A\&M University, College Station}\\
{\normalfont\small\textsuperscript{*}Equal contribution.}%
}

\iclrfinalcopy
\makeatletter
\patchcmd{\@maketitle}
  {Published as a conference paper at ICLR 2027}
  {Preprint}
  {}{\PackageError{vedje-preprint}{Unable to replace conference header}{}}
\patchcmd{\@maketitle}
  {\begin{tabular}[t]{l}\bf\rule{\z@}{24pt}\@author}
  {\begin{tabular}[t]{l}\bf\rule{\z@}{16pt}\@author}
  {}{\PackageError{vedje-preprint}{Unable to adjust author spacing}{}}
\patchcmd{\@maketitle}
  {\vskip 0.3in minus 0.1in}
  {\vskip 0pt}
  {}{\PackageError{vedje-preprint}{Unable to adjust title spacing}{}}
\makeatother
\hypersetup{%
  hidelinks,
  pdftitle={VEDJE: Video-Efficient Discriminative Joint Encoder for Scalable Video-Text Retrieval},
  pdfauthor={Shahaf Wagner, Gabriele Serussi, Dan Ben Ami, Tomer Galanti, Chaim Baskin},
  pdfsubject={Video-text retrieval},
  pdfkeywords={video retrieval, reranking, compact video caches}%
}

\begin{document}

\maketitle

\begin{abstract}
\edit{Finding the right video often requires distinguishing similar scenes in which different events occur. Joint matching improves retrieval, but processing rich video representations for each query is costly.
\architecture{} compresses features within sampled frames while keeping their representations separate in a reusable cache. Feature-change prediction supplies an auxiliary training signal that improves retrieval from the compressed cache without adding work at query time.
On MSR-VTT, MSVD, DiDeMo, and ActivityNet, \architecture{} improves R@1 over matched first-stage retrievers in both retrieval directions. On MSR-VTT, it reaches 59.8 text-to-video R@1 with a fine-tuned VideoCLIP-XL first stage. In the VideoPrism configuration, shrinking the per-video cache fourfold to 12\,KiB preserves text-to-video recall within 0.2 points.
These results show that accurate video search can operate on compact evidence, encoded once and reused as new queries arrive.\nobreak\hspace{0.5em}\githubcodelink}
\end{abstract}

\begin{figure}[H]
    \centering
    \includegraphics[width=\linewidth]{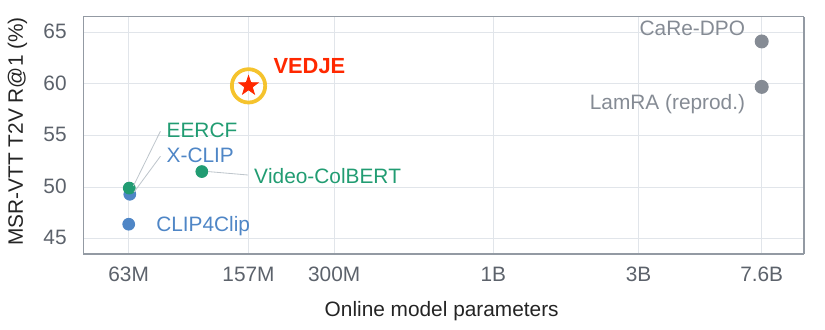}
    \caption{\edit{\textbf{MSR-VTT T2V R@1 (\%) versus online parameter count.} \architecture{} raises fine-tuned VideoCLIP-XL retrieval from 56.2 to 59.8 R@1 with a 33M-parameter cached reranker, bringing the total online footprint to approximately 157M parameters. Published systems retain their reported backbones and training protocols. Counts cover online text encoding and scoring; MLLM points count the base decoder, and Video-ColBERT counts its full text module. Appendix~\ref{app:comparison_context} details accounting and sources; Section~\ref{subsec:query_cost} reports complete-query latency.}}
    \label{fig:method_positioning}
\end{figure}

\section{Introduction}
\label{sec:introduction}

Video search at scale runs in two stages: a first stage retrieves plausible candidates over the full index, and a second stage verifies whether each candidate matches the query. \edit{This verification step must distinguish videos that share similar scenes but differ in the actions, state changes, or specific moments described by the query.}

\edit{Dual-encoder retrieval stages~\citep{luo2022clip4clip,ma2022x,wang_etal_2024_videoclip,wang2024internsideo2} index videos and queries independently to retrieve candidates cheaply, but compute similarity without joint token interaction. Joint rerankers, mature for image retrieval~\citep{BLIP,BLIP2}, restore cross-modal attention. For video, however, both stored features and joint scoring sequences scale with the sampled frames, while re-encoding candidates online incurs severe latency. Multimodal LLM (MLLM) rerankers~\citep{ko2025bidirectional,lee2025captioning} provide expressive joint matching, but place multi-billion-parameter decoders directly on the query path.}

One way to keep verification deployable is to remove the visual encoder from the query path. The image cached reranker \textsc{Edje}~\citep{taraday2026efficient} already shows how for static images: encode each image once offline, store a small set of compressed tokens, and rerank online with a lightweight joint encoder.

\edit{A video cache must be built before the queries it will serve are known. It must therefore retain evidence for descriptions of different objects, actions, and moments, while remaining small enough to store and score at scale. \architecture{} addresses this problem by compressing features within sampled frames and retaining separate representations across sampled times. A lightweight joint encoder then reads these representations alongside the query. Feature-change supervision improves retrieval from the compressed tokens without adding computation to the query path.}

\edit{Offline, a frozen visual encoder processes each video once, and a shared compressor extracts a few learned tokens per sampled frame. At query time, a 33M-parameter cross-encoder scores the cached tokens alongside the query text and first-stage score. The token budget determines both per-video storage and the visual sequence length presented to the online scorer.}

\edit{Across MSR-VTT~\citep{Xu_2016_CVPR}, MSVD~\citep{chen-dolan-2011-collecting}, DiDeMo~\citep{Hendricks_2017_ICCV}, and ActivityNet~\citep{Krishna_2017_ICCV}, \architecture{} consistently improves R@1 in both retrieval directions over matched first-stage retrievers. On MSR-VTT, it reaches 59.8 text-to-video R@1 with a fine-tuned VideoCLIP-XL first stage (Figure~\ref{fig:method_positioning}). In the VideoPrism configuration, allocating just one cached token per sampled frame reduces cache storage fourfold to 12\,KiB per video while preserving T2V R@1 within 0.2 points. Controlled comparisons evaluate cache structure, feature-change training, and joint scoring, while complete-query measurements benchmark practical retrieval latency.}

{\bf Contributions.\enspace} We make three contributions:
\begin{editblock}
\begin{enumerate}
    \item \textbf{Compact frame-indexed video caches:} Building on cached joint encoding, we compress information within sampled frames while retaining separate frame groups, giving the joint scorer access to evidence from across the sequence under a fixed storage budget.
    \item \textbf{Feature-change supervision for compression:} We train the compressed representation with future-delta feature prediction, improving retrieval under tight storage budgets without an inference-time predictor.
    \item \textbf{Accuracy and efficiency evaluation:} Across four benchmarks, \architecture{} improves R@1 in both directions over matched first stages. On MSR-VTT, it raises fine-tuned VideoCLIP-XL from 56.2 to 59.8 T2V R@1 with approximately 157M total online parameters. The VideoPrism configuration preserves T2V recall within 0.2 points with a fourfold-smaller 12\,KiB cache.
\end{enumerate}
\end{editblock}

\section{Related Work}
\label{sec:related_work}

\noindent\textbf{Dual encoders for video-text retrieval.\enspace}
\edit{CLIP-style contrastive retrieval stages embed videos and queries independently and score by inner product}~\citep{CLIP,luo2022clip4clip,wang_etal_2024_videoclip,wang2024internsideo2,zhao2024videoprism}; \edit{efficiency variants further compress stage 1~\citep{tian2024towards,tian2023teachclip}.} \edit{This dot product enables fast candidate retrieval, but forming the video embedding before the query arrives permanently fixes the visual representation without cross-modal interaction.}

\edit{\noindent\textbf{Late interaction and multi-vector matching.\enspace}
Late interaction stores richer unimodal token banks and aggregates similarities at query time~\citep{khattab2020Colbert,yao2022filip,Reddy_2025_CVPR}.} \edit{While multi-token caches preserve detail, the query and cached tokens never share a joint encoder.} \edit{EERCF~\citep{tian2024towards} instead pairs a trained temporal encoder with parameter-free text-gated feature aggregation.} \edit{We compare with EERCF on a shared CLIP ViT-B/16 architecture and separately evaluate \architecture{} on candidates from its released ViT-B/32 retriever.}

\noindent\textbf{Joint rerankers for vision-language retrieval.\enspace}
Joint rerankers restore pairwise inspection. For images, BLIP variants~\citep{BLIP,BLIP2} establish the recipe and \textsc{Edje}~\citep{taraday2026efficient} makes it deployable by caching compressed visual tokens. \edit{\textsc{Edje} also outlines prospective video extensions using temporal compression and positional encodings.} \edit{For video, CrossTVR~\citep{DAI2025111099} applies multi-grained cross-attention to selected frame- and video-level visual tokens, while multimodal LLM (MLLM) rerankers~\citep{ko2025bidirectional,lee2025captioning,liu2025lamra} place multi-billion-parameter decoders directly on the query path.} \edit{\architecture{} instead builds a compact, reusable cache that enables fine-grained joint attention with a lightweight 33M encoder.}

\edit{\noindent\textbf{Where \architecture{} sits.\enspace}
\architecture{} targets cached joint reranking for video; its closest neighbour is \textsc{Edje}'s image-caching recipe~\citep{taraday2026efficient}. 
\architecture{} develops and evaluates a video cache that compresses information within sampled frames while retaining separate representations across sampled times. Feature-change supervision improves retrieval from the compressed representation, and a lightweight joint encoder combines the cached evidence with the first-stage relevance estimate. Matched-budget comparisons show the benefit of retaining separate frame groups over pooling, while cache-size ablations measure how feature-change supervision helps at each budget. These results connect the organization and training of the cache to the retrieval accuracy it supports.}

\section{Method}
\label{sec:method}

\begin{figure}[H]
    \centering
    \includegraphics{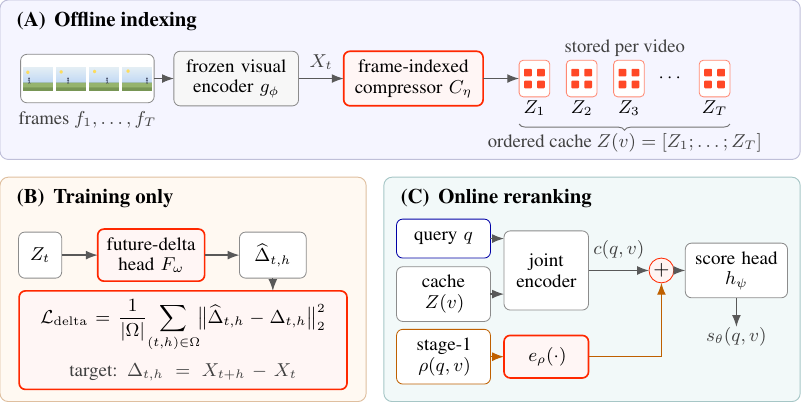}
    \caption{\edit{\architecture{} pipeline. \textbf{(A) Offline indexing:} compression of frame-indexed features yields the ordered cache $Z(v)=[Z_1;\dots;Z_T]$. \textbf{(B) Training only:} a future-delta head $F_\omega$ predicts feature changes from $Z_t$ and is discarded after training. \textbf{(C) Online reranking:} a 33M joint encoder processes $(q,Z(v))$; the embedded stage-1 score is added to its output before the nonlinear score head. Indexed videos require no query-time visual encoding.}}
    \label{fig:cache_story}
\end{figure}

\architecture{} replaces query-time video encoding with a \emph{cache-design problem}: the visual backbone runs once per video, offline, and writes a compact representation that a small joint encoder later reads at query time, alongside the query text and the scalar first-stage score. \edit{The cache must be small enough to ship at index scale while retaining evidence across the sampled sequence for later queries.} \edit{Figure~\ref{fig:cache_story} groups the three components that the rest of the section develops: (A)~the offline cache, (B)~the training-only feature-change supervision, and (C)~the online reranker that reads it.}

\Needspace{5\baselineskip}
\subsection{Setup and Notation}
\label{subsec:method_setup}

\edit{A first-stage retriever returns, for each text query $q$, a candidate set $\mathcal{C}(q)=\{v_i\}_{i=1}^{N_{\mathrm{c}}}$ together with a coarse score $\rho(q,v)\in\mathbb{R}$.} Each candidate is a video sampled into $T$ frames, $v=\{f_t\}_{t=1}^{T}$. \edit{A frozen visual encoder $g_\phi$ supplies frame-indexed patch features $X_t=[g_\phi(v)]_t\in\mathbb{R}^{P\times d_v}$, where $P$ is the number of spatial positions and $d_v$ is the visual feature dimension.} \edit{Each slice is indexed by sampled time and can include temporal context from the backbone.} \architecture{} writes a cache $Z(v)\in\mathbb{R}^{TM\times d_\ell}$ once per
video, storing $M$ tokens per frame. At query time, a reranker produces a score $s_\theta(q,v)\in\mathbb{R}$ from $q$, $Z(v)$, and $\rho(q,v)$, and sorting $\mathcal{C}(q)$ by $s_\theta$ gives the second-stage ranking. \edit{The product $TM$ controls both per-video storage and the online video sequence length.} \edit{The first-stage retriever and cache encoder serve separate roles within the corresponding backbone family.} \edit{The same pairwise score supports video-to-text retrieval over candidate captions for an indexed video; a new video query must first be encoded and cached.}

\subsection{\edit{A Frame-Indexed Cache across the Sampled Sequence}}
\label{subsec:compressor}

The cache $Z(v)$ must satisfy two opposing constraints. \edit{It must be small enough to ship per video at index scale, and it must retain evidence across the sampled sequence for a compact joint encoder to compare with future queries.} \edit{A single global clip vector uses little storage but collapses the sequence into one representation; storing every patch preserves separate frame features at a much higher cost.} \edit{\architecture{} resolves the tension by compressing each frame-indexed feature slice to a small token bank and placing that bank at its temporal position.}

\edit{A shared two-layer transformer decoder $C_\eta$ uses $M$ learned cache queries $Q_0\in\mathbb{R}^{M\times d_\ell}$ to attend to the projected patch tokens of each frame-indexed slice,}
\begin{editblock}
\begin{equation}
\label{eq:frame_compressor}
U_t=X_tW_{\mathrm{kv}},\qquad
Z_t=C_\eta(Q_0;U_t)\in\mathbb{R}^{M\times d_\ell},
\end{equation}
\end{editblock}
\edit{where $W_{\mathrm{kv}}\in\mathbb{R}^{d_v\times d_\ell}$ is the shared input projection for keys and values.} The stored cache is the temporal concatenation
\begin{equation}
\label{eq:ordered_cache}
Z(v) = [Z_1;\,\dots;\,Z_T] \in \mathbb{R}^{TM\times d_\ell}.
\end{equation}
This construction has two consequences. Per-frame compression bounds storage by $T\times M$ cached tokens and is independent of the patch count $P$ of $g_\phi$. \edit{Ordered concatenation retains a separate token group for each sampled time even when $M$ is small.} \edit{The cache-structure ablation in Section~\ref{subsec:cache_structure} contrasts this design with pooled multi-frame caches under matched cache size.}

\edit{The token budget also gives an explicit storage budget. Each BF16 element occupies two bytes, so the cached tensor payload is}
\begin{editblock}
\[
B_{\mathrm{cache}}=2TMd_\ell\ \text{bytes per video}.
\]
\end{editblock}
\edit{With $T{=}16$ and $d_\ell{=}384$, four tokens per frame yield 48\,KiB, while one token per frame yields 12\,KiB. Reducing $M$ retains the same sampled times but compresses each feature slice more strongly. Appendix Table~\ref{tab:app_visual_payload} compares these payloads with the full frame-and-patch representation at the same precision.}

\subsection{\edit{Feature-Change Supervision for a Tight Cache}}
\label{subsec:delta_fullgrid}

The cache is query independent: once $Z(v)$ is written, every future query about $v$ (about objects, actions, or temporal relations) reads the same tokens. \edit{Feature-change supervision adds a training target at each patch position by predicting differences between the frozen features of sampled frames.}

\edit{A lightweight, training-only predictor $F_\omega$ learns to estimate these changes from $Z_t$,}
\begin{equation}
\label{eq:delta_pred}
\widehat{\Delta}_t = F_\omega(Z_t)\in\mathbb{R}^{|\mathcal{H}|\times P\times d_v},
\qquad
\Delta_{t,h} = X_{t+h}-X_t \in \mathbb{R}^{P\times d_v}.
\end{equation}
\edit{Subtracting $X_t$ cancels feature components that are identical at both sampled times.} \edit{The target describes change within the encoded clip: both feature slices can already include temporal context from the frozen backbone.} \edit{The predictor outputs one prediction per horizon and spatial position, so all $P$ patch targets must be estimated from the same $M$ cached tokens.} \edit{These targets compare feature-grid positions across time; they supervise feature changes without requiring motion annotations or enlarging the cache.}

The loss averages the squared $\ell_2$ distance to the frozen target over the valid time-horizon pairs $\Omega=\{(t,h)\mid 1\le t\le T,\;h\in\mathcal{H},\;t+h\le T\}$,
\begin{equation}
\label{eq:delta_loss}
\deltaloss
=
\frac{1}{|\Omega|}
\sum_{(t,h)\in\Omega}
\big\|\widehat{\Delta}_{t,h} - \Delta_{t,h}\big\|_2^2,
\end{equation}
where $\|\cdot\|_2^2$ is the per-element-averaged squared $\ell_2$ norm over patches and the visual feature dimension; \edit{the frozen $g_\phi$ makes $\Delta_{t,h}$ a fixed regression target, so gradients update the predictor and compressor, not the visual backbone.} \edit{After training, $F_\omega$ is discarded, so the auxiliary objective adds no computation at retrieval.} \edit{Section~\ref{subsec:future_delta_compression} measures its effect across cache budgets.}

\subsection{Cached Joint Reranking with a Residual Prior}
\label{subsec:reranker}

\edit{The reranker adds a compact transformer encoder to the query path.} \edit{It tokenizes $q$, concatenates the text tokens with the cached video tokens $Z(v)$, and produces a pooled pairwise representation $c(q,v)\in\mathbb{R}^{d_\ell}$.} \edit{Its own learned positional embedding covers the full input $[\text{tokens}(q);\,Z(v)]$, making the temporal layout fixed at indexing time (Eq.~\ref{eq:ordered_cache}) available to self-attention.} \edit{Because $c(q,v)$ is computed after the query and the cache see each other, the pairwise representation can depend on their joint content.}

The first-stage score $\rho(q,v)$ carries different information. \edit{It supplies the coarse relevance estimate that produced the candidate list.} \edit{\architecture{} includes this estimate as a residual prior by embedding the scalar score and adding it to the joint representation,}
\begin{editblock}
\begin{equation}
\label{eq:residual_score}
\tilde{c}(q,v) = c(q,v) + e_\rho\!\big(\rho(q,v)\big),
\qquad
s_\theta(q,v) = h_\psi\!\left(\tilde{c}(q,v)\right),
\end{equation}
\end{editblock}
\edit{where $e_\rho:\mathbb{R}\to\mathbb{R}^{d_\ell}$ and $h_\psi:\mathbb{R}^{d_\ell}\to\mathbb{R}$ are two-layer MLPs with GELU activation.} \edit{The residual addition occurs in representation space, followed by the nonlinear score head.} \edit{Section~\ref{subsec:stage1_prior} compares the full model with independently trained single-stream controls under fixed candidates.}

\subsection{\edit{Training and Deployment}}
\label{subsec:training_objectives}

\edit{\architecture{} jointly trains the compressor, joint encoder, prior embedding, score head, and auxiliary training heads with four objectives,}
\begin{equation}
\label{eq:total_loss}
\mathcal{L}
=
\mathcal{L}_{\mathrm{vtm}}
+\mathcal{L}_{\mathrm{vtc}}
+\mathcal{L}_{\mathrm{mlm}}
+\deltaloss.
\end{equation}
\edit{All four terms enter at equal weight, and the loss-stack ablation in Appendix~\ref{app:loss_stack_ablation} evaluates their contributions at this fixed weighting.}
\edit{The matching loss $\mathcal{L}_{\mathrm{vtm}}$ supervises the pairwise score $s_\theta(q,v)$ with cross-entropy over matched and negative pairs.} \edit{The contrastive loss $\mathcal{L}_{\mathrm{vtc}}$ aligns projected text CLS representations with frozen video embeddings from the corresponding backbone through symmetric in-batch InfoNCE.} \edit{The masked-language loss $\mathcal{L}_{\mathrm{mlm}}$ predicts masked caption tokens from the surrounding text with cached video tokens in context.} \edit{\deltaloss{} supplies the frame-level feature-change targets in Eq.~\ref{eq:delta_loss}.} \edit{Appendix~\ref{app:training_protocol} gives the loss definitions and optimization schedule.}

\edit{After training, the auxiliary heads are discarded.} \edit{Each text query requires first-stage text encoding and search, cache access, and the compact joint reranker with $e_\rho$ and $h_\psi$.} Algorithm~\ref{alg:vedje} in Appendix~\ref{app:algorithm} writes out indexing and reranking. \edit{Updating only the joint reranker permits cache reuse; changing the visual backbone or compressor requires rebuilding the cache.} \edit{Section~\ref{sec:experiments} evaluates per-video storage and the complete query cost.}

\section{Experiments}
\label{sec:experiments}

\edit{The experiments evaluate retrieval quality, cache design, and complete-query cost. Matched controls examine how frame grouping, feature-change supervision, and joint scoring contribute.}

\subsection{Experimental Setup}
\label{subsec:exp_protocol}

\architecture{} is evaluated on MSR-VTT~\citep{Xu_2016_CVPR}, MSVD~\citep{chen-dolan-2011-collecting}, DiDeMo~\citep{Hendricks_2017_ICCV}, and ActivityNet~\citep{Krishna_2017_ICCV}, using standard splits and Recall@$K$ in both retrieval directions: text-to-video (T2V) and video-to-text (V2T). \edit{Recall values are percentages; absolute differences are percentage points.}
\edit{A first-stage retriever returns the top-$N_{\mathrm{c}}$ candidates (default $N_{\mathrm{c}}{=}20$), which \architecture{} then reranks from cached video tokens.} \edit{The default \architecture{} configuration uses a 64-token BF16 cache, the VideoPrism prior~\citep{zhao2024videoprism}, and prediction horizon $\mathcal{H}=\{3\}$.}
\edit{A variant is written \vedjevariant{S1}{64}, where the subscript identifies the backbone used for retrieval and cached features: VideoPrism (VP), VideoCLIP-XL (VCLIP)~\citep{wang_etal_2024_videoclip}, or Perception Encoder (PE)~\citep{bolya2025perceptionencoder}. Non-default cache sizes are appended to the variant name.} \edit{Parenthesized gains report absolute R@1 improvements over the matched first-stage retriever.}
\edit{The default joint reranker is MiniLM-L12-uncased~\citep{wang2020minilm_3f5ee243}, a compact 12-layer language model with $\sim$33M parameters.} \edit{It jointly scores cached video tokens alongside query text for T2V, and alongside candidate captions for an indexed video for V2T.}
\edit{Each \architecture{} variant constructs its cache from the corresponding visual backbone and uses that backbone\textquotesingle s video embeddings as contrastive training targets (Appendix Table~\ref{tab:backbone_sources}). The cache retains compressed patch features for joint scoring, whereas the contrastive objective uses the corresponding video-level embedding as a training target. During reranker training, the feature encoder remains frozen while the compressor and joint scorer are optimized. ZS and FT describe the first-stage retriever; the reranker is trained in both settings. The CLIP control uses CLIP features and contrastive targets throughout (Table~\ref{tab:controlled_comparisons}).} \edit{The video input uses 16 uniformly sampled frames. The joint scorer accepts up to 64 text tokens including special tokens; Appendix~\ref{app:arch_details} gives the architecture and input budgets.}

\edit{Cache ablations report means and population standard deviations over three training seeds. Values are rounded to one decimal, with a minimum displayed SD of 0.1.} \edit{Gain standard deviations describe variability across retrainings. The separate four-run auxiliary check reports the standard deviation of paired gains and is identified independently of the three-seed cache ablations. Appendix~\ref{app:details} gives the statistical protocol.}

\subsection{\edit{Q1: Does Reranking Improve Retrieval Across Settings?}}
\label{subsec:competitors}
\label{subsec:controlled_comparisons}

\edit{\architecture{} improves R@1 over each first-stage retriever in \hyperref[tab:msrvtt_matched_panel]{Table~\ref*{tab:msrvtt_overview}b}, including fine-tuned VideoCLIP-XL. In that setting, reranking raises T2V R@1 from 56.2 to 59.8 and V2T R@1 from 55.1 to 58.8. Candidates and first-stage scores are regenerated for the fine-tuned model, and the reranker is retrained using VideoCLIP-XL cache features. The gains therefore remain after first-stage fine-tuning.}

\begin{table}[t]
\centering
\caption{\edit{\textbf{MSR-VTT retrieval accuracy.} R@1 is in percent; gains are percentage points. (a) Source protocols are retained; UMT-L and InternVideo2 report fine-tuned results; UMT-L uses 25M pretraining pairs, and CLIP4Clip reports the best variant per direction. LamRA uses the VideoChat-Flash reproduction of \citet{lee2025captioning}. (b) Matched reranking uses a 64-token cache from the corresponding backbone; fine-tuned first-stage results are our reproduction. ZS/FT denote zero-shot/fine-tuned first stages; dashes denote unreported results. Full results and uncertainty appear in Appendix~\ref{app:full_retrieval_results}.}}
\label{tab:msrvtt_overview}
\begin{editblock}
\phantomsection\label{tab:msrvtt_published_panel}
\begin{tabularx}{\linewidth}{@{}X rr@{}}
\toprule
\multicolumn{3}{@{}l}{\textbf{(a) Published systems}} \\
Method & T2V R@1 & V2T R@1 \\
\midrule
CLIP4Clip B/32~\citep{luo2022clip4clip} & 44.5 & 43.1 \\
X-CLIP B/16~\citep{ma2022x} & 49.3 & 48.9 \\
Video-ColBERT (SigLIP-B/16)~\citep{Reddy_2025_CVPR} & 51.5 & -- \\
CrossTVR-Large~\citep{DAI2025111099} & 54.0 & 51.3 \\
UMT-L (ViT-L/16)~\citep{li2023unmasked} & 58.8 & 58.6 \\
LamRA~\citep{liu2025lamra} & 59.7 & 60.7 \\
InternVideo2 (stage 2, 6B)~\citep{wang2024internsideo2} & 62.8 & 60.2 \\
CaRe-DPO~\citep{lee2025captioning} & 64.1 & 63.8 \\
\bottomrule
\end{tabularx}
\par\smallskip
\phantomsection\label{tab:msrvtt_matched_panel}
\begingroup
\setlength{\tabcolsep}{3.5pt}
\begin{tabularx}{\linewidth}{@{}l >{\raggedleft\arraybackslash}X >{\columncolor{ourmethodrow}\raggedleft\arraybackslash}X >{\raggedleft\arraybackslash}X >{\raggedleft\arraybackslash}X >{\columncolor{ourmethodrow}\raggedleft\arraybackslash}X >{\raggedleft\arraybackslash}X@{}}
\multicolumn{7}{@{}l}{\textbf{(b) Matched first-stage comparisons}} \\
& \multicolumn{3}{c}{\textbf{Text-to-video}} & \multicolumn{3}{c@{}}{\textbf{Video-to-text}} \\
\cmidrule(lr){2-4}\cmidrule(l){5-7}
First-stage retriever & Stage 1 & VEDJE & Gain & Stage 1 & VEDJE & Gain \\
\midrule
VideoPrism & 50.1 & \textbf{54.6} & +4.5 & 49.8 & \textbf{54.6} & +4.8 \\
PE-Core-B & 47.6 & \textbf{53.9} & +6.3 & 47.3 & \textbf{53.9} & +6.6 \\
VideoCLIP-XL (ZS) & 50.1 & \textbf{56.5} & +6.4 & 49.9 & \textbf{57.1} & +7.2 \\
VideoCLIP-XL (FT) & 56.2 & \textbf{59.8} & +3.6 & 55.1 & \textbf{58.8} & +3.7 \\
\bottomrule
\end{tabularx}
\endgroup
\end{editblock}
\end{table}

\edit{The cross-dataset evaluation extends this comparison to MSR-VTT, MSVD, DiDeMo, and ActivityNet (Appendix Table~\ref{tab:cross_dataset_r1}). With either VideoPrism or zero-shot VideoCLIP-XL as the first stage, \architecture{} improves R@1 in both retrieval directions on all four datasets. Published systems in \hyperref[tab:msrvtt_published_panel]{Table~\ref*{tab:msrvtt_overview}a} provide broader context under their reported backbones and training protocols; Table~\ref{tab:controlled_comparisons} examines performance with a shared CLIP ViT-B/16 architecture.}

\begin{table}[t]
\centering
\caption{\edit{\textbf{Shared CLIP ViT-B/16 architecture on MSR-VTT.} R@1 (\%). Published baselines fine-tune CLIP. \architecture{} keeps CLIP fixed and trains its compressor and reranker, using CLIP features and contrastive targets. CLIP4Clip values are the reproduction of \citet{tian2024towards}.}}
\label{tab:controlled_comparisons}
\begin{editblock}
\begin{tabularx}{\linewidth}{X l rr}
\toprule
Method & Backbone treatment & T2V R@1 & V2T R@1 \\
\midrule
CLIP4Clip~\citep{luo2022clip4clip} & fine-tuned & 46.4 & 45.4 \\
X-CLIP~\citep{ma2022x} & fine-tuned & 49.3 & 48.9 \\
EERCF~\citep{tian2024towards} & fine-tuned & 49.9 & 47.8 \\
\oursrow VEDJE-CLIP & frozen & 49.8 & 49.6 \\
\bottomrule
\end{tabularx}
\end{editblock}
\end{table}

\edit{Using CLIP ViT-B/16 for both cached visual features and contrastive training targets, \architecture{} reaches 49.8/49.6 T2V/V2T R@1 (Table~\ref{tab:controlled_comparisons}). EERCF reports 49.9/47.8 on the same backbone architecture. \architecture{} achieves comparable T2V recall and 1.8 points higher V2T recall while keeping the visual backbone frozen.} \edit{A further control uses EERCF's released ViT-B/32 candidate generator ($N_{\mathrm{c}}{=}50$): we reproduce 42.8 T2V R@1 before reranking and reach 49.1 with \architecture{}. EERCF reports 47.8 under its published protocol (Appendix~\ref{app:controlled_comparisons}).} \edit{Figure~\ref{fig:method_positioning} and Appendix~\ref{app:comparison_context} place these findings in the broader model-footprint comparison.}

\edit{These controlled comparisons isolate reranking gains over candidate retrieval; Section~\ref{subsec:stage1_prior} tests scoring rules with cached tokens held fixed.}

\subsection{Q2: How Small Can the Cache Become?}
\label{subsec:token_budget}

\edit{The default 64-token cache occupies 48\,KiB per video. Table~\ref{tab:token_budget_storage} reduces this budget to one token per sampled frame, producing a 16-token cache of 12\,KiB, and evaluates both budgets with and without feature-change supervision. The sampled frames and visual backbone remain fixed.}

\begin{table}[t]
\centering
\caption{\edit{\textbf{Cache size and feature-change supervision.} MSR-VTT R@1 (\%) is mean $\pm$ standard deviation over three seeds. Values are rounded to one decimal, with a minimum displayed SD of 0.1. Storage is BF16 tensor payload per video; gains are percentage points.}}
\label{tab:token_budget_storage}
\label{tab:future_delta_token_budget}
\begin{editblock}
\begin{tabularx}{\linewidth}{r r c *{2}{>{\raggedleft\arraybackslash}X} r}
\toprule
Tokens & KiB/video & Direction & Without $\deltaloss$ & With $\deltaloss$ & Gain \\
\midrule
64 & 48 & T2V & $54.0\pm0.1$ & $\mathbf{54.6}\pm0.1$ & +0.6 \\
64 & 48 & V2T & $53.6\pm0.1$ & $\mathbf{54.6}\pm0.1$ & +1.0 \\
\oursrow 16 & 12 & T2V & $52.5\pm0.1$ & $\mathbf{54.4}\pm0.1$ & +1.9 \\
\oursrow 16 & 12 & V2T & $52.6\pm0.1$ & $\mathbf{53.1}\pm0.1$ & +0.5 \\
\bottomrule
\end{tabularx}
\end{editblock}
\end{table}

\edit{With feature-change supervision, reducing storage fourfold changes T2V R@1 from 54.6 to 54.4, while V2T R@1 falls from 54.6 to 53.1. The compact configuration therefore offers a favorable storage trade-off for T2V, with a larger accuracy cost for V2T. The comparison measures the effect of compression relative to the 48\,KiB cache. Appendix~\ref{app:latency_io_protocol} reports the corresponding tensor payloads, and Appendix~\ref{app:token_precision_details} examines cache precision.} \edit{The 48\,KiB cache is 128.5 times smaller than the full frame-and-patch representation of the same VideoPrism features (Appendix Table~\ref{tab:app_visual_payload}).}

\subsection{Q3: Which Cache Structure Carries the Gain?}
\label{subsec:cache_structure}

\edit{At a fixed storage budget, should the cache pool evidence across frames or keep frame groups separate? Table~\ref{tab:cache_structure} compares mean pooling, attention pooling, and separate frame groups using the same VideoPrism candidates, 33M reranker, and 64-token cache.}

\begin{table}[t]
\centering
\caption{\edit{\textbf{Cache structure at a fixed storage budget.} MSR-VTT R@1 (\%) is mean $\pm$ standard deviation over three training seeds. The multiframe variants share VideoPrism candidates, the first-stage score input, the 33M reranker, and a 64-token BF16 cache. Pooling combines evidence across frames; \architecture{} retains separate frame groups.}}
\label{tab:cache_structure}
\begin{editblock}
\begin{tabularx}{\linewidth}{X c rr}
\toprule
Cache construction & $\deltaloss$ & T2V R@1 & V2T R@1 \\
\midrule
FrameSet-EDJE: mean pooling & No & $51.7\pm0.2$ & $51.6\pm0.2$ \\
FrameSet-EDJE: attention pooling & No & $52.0\pm0.2$ & $51.6\pm0.2$ \\
VEDJE: separate frame groups & No & $54.0\pm0.1$ & $53.6\pm0.1$ \\
\oursrow VEDJE: separate frame groups & Yes & $\mathbf{54.6}\pm0.1$ & $\mathbf{54.6}\pm0.1$ \\
\bottomrule
\end{tabularx}
\end{editblock}
\end{table}

\edit{Before auxiliary training, retaining separate frame groups improves R@1 by 2.0 points in both directions over attention pooling. Feature-change supervision raises recall further to 54.6 in both directions. The matched comparison shows that the organization of cached evidence affects retrieval even when storage remains fixed.} \edit{\hyperref[app:single_frame_reference]{Appendix~\ref*{app:controlled_comparisons}} reports the single-frame EDJE-style reference separately.}

\subsection{\edit{Q4: Where Does Feature-Change Supervision Help?}}
\label{subsec:future_delta_compression}
\label{subsec:auxiliary_design}

\edit{Feature-change supervision improves R@1 at both tested cache budgets (Table~\ref{tab:token_budget_storage}). Its measured T2V gain increases from 0.6 points with 64 tokens to 1.9 points with 16 tokens, showing its value when each sampled frame receives only one cached token. V2T follows a different pattern, with the larger gain at 64 tokens. A separate four-run check at that budget reports positive gains in every run (\hyperref[app:four_run_auxiliary]{Appendix~\ref*{app:controlled_comparisons}}).}

\edit{At 64 tokens, current-frame reconstruction reaches 54.2 T2V and 53.9 V2T R@1, compared with 54.6 in both directions for the delta target (Appendix Table~\ref{tab:auxiliary_targets}). This comparison favors feature-change supervision over the tested reconstruction target.}

\edit{The appendix examines target formulation (Appendix~\ref{app:auxiliary_targets}) and prediction offset (Appendix~\ref{app:horizon_sweep}). The matched comparisons in Table~\ref{tab:token_budget_storage} establish the retrieval gains from adding the objective at each cache budget.}

\subsection{\edit{Q5: What Do Joint Scoring and the Prior Contribute?}}
\label{subsec:stage1_prior}

\edit{Table~\ref{tab:main_same_cache_scoring} holds the cached representation and candidate pool fixed to compare how different scorers use the same evidence. The joint encoder reaches 54.6 T2V R@1, compared with 51.9 for trained MaxSim under the same data, schedule, and first-stage score input.}

\begin{table}[t]
\centering
\caption{\edit{\textbf{Scoring the same cache on MSR-VTT.} T2V R@1 (\%) with VP stage 1 and 20 candidates. Reranking rows share the 64-token cache and candidate pool; trained MaxSim uses the same data and schedule. Appendix~\ref{app:controlled_comparisons} gives the protocols.}}
\label{tab:main_same_cache_scoring}
\begin{editblock}
\begin{tabularx}{\linewidth}{X r}
\toprule
Scoring rule & T2V R@1 \\
\midrule
Stage 1: VideoPrism & 50.1 \\
\midrule
Trained MaxSim with residual prior & 51.9 \\
\oursrow Joint encoder with residual prior (VEDJE) & \textbf{54.6} \\
\bottomrule
\end{tabularx}
\end{editblock}
\end{table}

\edit{The first-stage score and joint features also contribute complementary information. Independently trained prior-only and joint-only models reach 50.1 and 43.8 T2V R@1, while the full model reaches 54.6 (Appendix Table~\ref{tab:stage1_prior_controls}). Together, these comparisons support using joint matching to refine the relevance estimate supplied by the first stage.}

\Needspace{4\baselineskip}
\subsection{\edit{Q6: What Does a Complete Query Cost?}}
\label{subsec:query_cost}

\edit{Table~\ref{tab:app_online_cost} measures the complete VideoPrism query pipeline on an NVIDIA L40S, including text encoding, index search, cache reads, transfer to the GPU, and reranking. With 20 candidates, the 64-token configuration takes 5.79\,ms, compared with approximately 4.3\,ms for first-stage retrieval alone.} \edit{This configuration has approximately 143M total online parameters, including the 33M reranker (Appendix~\ref{app:comparison_context}).}

\begin{table}[t]
\centering
\caption{\edit{\textbf{Complete-query cost at two cache sizes.} MSR-VTT T2V with VP stage 1, host-cache I/O, one L40S, and query batch size 1. Latency is the median of 200 runs after 30 warmups. R@1 at 20 candidates is a three-seed mean; other recalls are single-run evaluations.}}
\label{tab:app_online_cost}
\begin{editblock}
\begin{tabularx}{\linewidth}{l *{4}{>{\raggedleft\arraybackslash}X}}
\toprule
& \multicolumn{2}{c}{Latency (ms)} & \multicolumn{2}{c@{}}{T2V R@1 (\%)} \\
\cmidrule(lr){2-3}\cmidrule(l){4-5}
Candidates $N_{\mathrm{c}}$ & 64 tokens & 16 tokens & 64 tokens & 16 tokens \\
\midrule
\oursrow 20 (default) & 5.79 & 5.82 & 54.6 & 54.4 \\
50 & 6.21 & 6.12 & 53.9 & 54.7 \\
100 & 7.83 & 6.33 & 54.4 & 54.6 \\
200 & 10.24 & 6.88 & 54.3 & 54.7 \\
\bottomrule
\end{tabularx}
\end{editblock}
\end{table}

\edit{The two cache sizes have similar complete-query latency at 20 candidates. Text encoding and index search are shared by both configurations, so reducing the visual sequence affects only part of the complete-query cost. Their difference grows as the scorer processes more candidates: at 100, the 16-token cache reduces latency from 7.83 to 6.33\,ms. Compression also reduces the video payload read for each query. The appendix separates cache-residency effects from device-resident scoring (Appendix~\ref{app:latency_io_protocol}) and reports candidate-count sensitivity (Appendix~\ref{app:k_sensitivity}).}

\edit{Cache residency directly affects retrieval latency: at 100 candidates, the 64-token model requires 7.8\,ms under warm NVMe reads and 11.1\,ms under cold reads.} \edit{At 20 candidates, the 64- and 16-token caches contribute 0.983 and 0.246\,MB of video payload, respectively.} \edit{Only the retrieved candidates' caches are read, so this part of query traffic scales with candidate count and cache size rather than corpus size.} \edit{The first-stage search index is separate from these payloads.} \edit{Appendix~\ref{app:latency_io_protocol} breaks down cache residency and isolates complete-query latency from device-resident reranking.} \edit{Offline visual encoding and cache construction are paid once during indexing and amortized across all subsequent queries.}

\section{Conclusion}
\label{sec:conclusion}

\edit{A reusable video cache must retain useful evidence before its future queries are known. \architecture{} addresses this constraint by compressing features within sampled frames while keeping evidence from different times separately accessible. Matched comparisons show the value of this organization, and feature-change supervision improves retrieval from the compressed representation. Across four benchmarks, the resulting cache supports gains over first-stage retrieval, including a fine-tuned first stage. These findings establish a practical route to joint video search in which compact evidence is computed once and reused for later queries.}

\edit{\paragraph{Limitations and future work.}
The cache is \emph{query-blind}: $Z(v)$ is written once and serves every later query about $v$, so selectively reading or expanding the cache at query time is a natural extension.} \edit{The cache also depends on a \emph{fixed visual backbone and compressor}: updating either requires re-indexing the corpus.} \edit{The first and second stages are not currently trained jointly; closing that loop is another direction.} \edit{Finally, the four benchmarks here (MSR-VTT, MSVD, DiDeMo, ActivityNet) use English captions and web-sourced videos. These evaluations span different video lengths, but do not establish performance on multilingual video retrieval.}

\clearpage
\subsection*{\rev{AI use statement}}
\begin{editblock}
Generative AI was used to assist in writing the manuscript and in writing code, as well as in organizing material and checking consistency. All empirical results were supplied by the authors. Everything has been thoroughly checked and verified by the authors, who take full responsibility for the entire content of this submission.
\end{editblock}

\subsection*{\rev{Ethics statement}}
\begin{revisionblock}
This work collects no new data and involves no crowdsourcing, human-subject study, biometric processing, facial recognition, or identity annotation. Experiments use the established MSR-VTT, MSVD, DiDeMo, and ActivityNet benchmarks under their original release terms. These web-video datasets can nevertheless contain identifiable people, and their historical collection and consent documentation may not satisfy current expectations for every downstream use. Making semantic video retrieval cheaper can also increase cross-collection re-identification or surveillance risk when combined with external identity systems. Deployments should therefore require lawful and authorized data use, role-based access, purpose limitation, query logging and auditing, restrictions on identity-oriented searches, retention limits, and human review for surveillance or biometric applications. Automated query moderation can complement these controls.
\end{revisionblock}

\subsection*{\rev{Reproducibility statement}}
\begin{revisionblock}
Algorithm~\ref{alg:vedje} specifies indexing and reranking; Appendices~\ref{app:arch_details}--\ref{app:compute_code_release} give architecture, optimization, cache quantization, compute, code, and asset details. Appendix~\ref{app:latency_io_protocol} defines end-to-end latency and I/O measurement, Appendix~\ref{app:controlled_comparisons} documents matched controls, and Appendix~\ref{app:k_sensitivity} reports candidate-pool sensitivity. Appendix~\ref{app:details} describes how recall values and gains are summarized. \edit{Code:~\githubcodelink.}
\end{revisionblock}


\bibliography{references}

@inproceedings{tian2023teachclip,
  author = {Tian, Kaibin and Zhao, Ruixiang and Xin, Zijie and Lan, Bangxiang and Li, Xirong},
  title = {Holistic Features are Almost Sufficient for Text-to-Video Retrieval},
  booktitle = {Proceedings of the IEEE/CVF Conference on Computer Vision and Pattern Recognition (CVPR)},
  year = {2024},
  month = {June},
  pages = {17138--17147},
  doi = {10.1109/cvpr52733.2024.01622},
  url = {https://doi.org/10.1109/cvpr52733.2024.01622},
}

@inproceedings{tian2024towards,
  author = {Tian, Kaibin and Cheng, Yanhua and Liu, Yi and Hou, Xinglin and Chen, Quan and Li, Han},
  title = {Towards Efficient and Effective Text-to-Video Retrieval with Coarse-to-Fine Visual Representation Learning},
  booktitle = {Proceedings of the AAAI Conference on Artificial Intelligence},
  volume = {38},
  pages = {5207--5214},
  year = {2024},
  doi = {10.1609/aaai.v38i6.28327},
  url = {https://doi.org/10.1609/aaai.v38i6.28327},
}

@article{DAI2025111099,
  author = {Dai, Zuozhuo and Cheng, Kaihui and Shao, Fangtao and Dong, Zilong and Zhu, Siyu},
  title = {Text--video retrieval re-ranking via multi-grained cross attention and frozen image encoders},
  journal = {Pattern Recognition},
  volume = {159},
  pages = {111099},
  year = {2025},
  doi = {10.1016/j.patcog.2024.111099},
  url = {https://doi.org/10.1016/j.patcog.2024.111099},
  publisher = {Elsevier},
}

@inproceedings{ko2025bidirectional,
    author    = {Ko, Dohwan and Lee, Ji Soo and Choi, Minhyuk and Meng, Zihang and Kim, Hyunwoo J.},
    title     = {Bidirectional Likelihood Estimation with Multi-Modal Large Language Models for Text-Video Retrieval},
    booktitle = {Proceedings of the IEEE/CVF International Conference on Computer Vision (ICCV)},
    month     = {October},
    year      = {2025},
    pages     = {22263-22273},
    url = {https://doi.org/10.1109/iccv51701.2025.02067},
}

@inproceedings{wang2024internsideo2,
  author = {Wang, Yi and Li, Kunchang and Li, Xinhao and Yu, Jiashuo and He, Yinan and Chen, Guo and Pei, Baoqi and Zheng, Rongkun and Wang, Zun and Shi, Yansong and Jiang, Tianxiang and Li, Songze and Xu, Jilan and Zhang, Hongjie and Huang, Yifei and Qiao, Yu and Wang, Yali and Wang, Limin},
  title = {{InternVideo2}: Scaling Foundation Models for Multimodal Video Understanding},
  booktitle = {Computer Vision -- ECCV 2024},
  pages = {396--416},
  year = {2024},
  publisher = {Springer Nature Switzerland},
  doi = {10.1007/978-3-031-73013-9_23},
  url = {https://doi.org/10.1007/978-3-031-73013-9_23},
}

@inproceedings{lee2025captioning,
    title = "Captioning for Text-Video Retrieval via Dual-Group Direct Preference Optimization",
    author = "Lee, Ji Soo  and
      Ko, Byungoh  and
      Cho, Jaewon  and
      Lee, Howoong  and
      Byun, Jaewoon  and
      Kim, Hyunwoo J.",
    booktitle = "Findings of the Association for Computational Linguistics: EMNLP 2025",
    month = nov,
    year = "2025",
    address = "Suzhou, China",
    publisher = "Association for Computational Linguistics",
    url = "https://aclanthology.org/2025.findings-emnlp.869/",
    doi = "10.18653/v1/2025.findings-emnlp.869",
    pages = "16022--16039",
    ISBN = "979-8-89176-335-7",
}

@inproceedings{liu2025lamra,
      author    = {Liu, Yikun and Zhang, Yajie and Cai, Jiayin and Jiang, Xiaolong and Hu, Yao and Yao, Jiangchao and Wang, Yanfeng and Xie, Weidi},
    title     = {{LamRA}: Large Multimodal Model as Your Advanced Retrieval Assistant},
    booktitle = {Proceedings of the IEEE/CVF Conference on Computer Vision and Pattern Recognition (CVPR)},
    month     = {June},
    year      = {2025},
    pages     = {4015-4025},
  doi = {10.1109/cvpr52734.2025.00380},
  url = {https://doi.org/10.1109/cvpr52734.2025.00380},
}

@inproceedings{zhao2024videoprism,
  title = 	 {{V}ideo{P}rism: A Foundational Visual Encoder for Video Understanding},
  author =       {Zhao, Long and Gundavarapu, Nitesh Bharadwaj and Yuan, Liangzhe and Zhou, Hao and Yan, Shen and Sun, Jennifer J. and Friedman, Luke and Qian, Rui and Weyand, Tobias and Zhao, Yue and Hornung, Rachel and Schroff, Florian and Yang, Ming-Hsuan and Ross, David A and Wang, Huisheng and Adam, Hartwig and Sirotenko, Mikhail and Liu, Ting and Gong, Boqing},
  booktitle = 	 {Proceedings of the 41st International Conference on Machine Learning},
  pages = 	 {60785--60811},
  year = 	 {2024},
  editor = 	 {Salakhutdinov, Ruslan and Kolter, Zico and Heller, Katherine and Weller, Adrian and Oliver, Nuria and Scarlett, Jonathan and Berkenkamp, Felix},
  volume = 	 {235},
  series = 	 {Proceedings of Machine Learning Research},
  month = 	 {21--27 Jul},
  publisher =    {PMLR},
  url = 	 {https://proceedings.mlr.press/v235/zhao24f.html},
}

@inproceedings{bolya2025perceptionencoder,
 author = {Bolya, Daniel and Huang, Po-Yao and Sun, Peize and Cho, Jang Hyun and Madotto, Andrea and Wei, Chen and Ma, Tengyu and Zhi, Jiale and Rajasegaran, Jathushan and Bangalath, Hanoona and Wang, Junke and Monteiro, Marco and Xu, Hu and Dong, Shiyu and Ravi, Nikhila and Li, Shang-Wen and Dollar, Piotr and Feichtenhofer, Christoph},
 booktitle = {Advances in Neural Information Processing Systems},
 doi = {10.52202/085713-2036},
 editor = {D. Belgrave and C. Zhang and H. Lin and R. Pascanu and P. Koniusz and M. Ghassemi and N. Chen},
 pages = {60884--60937},
 publisher = {Curran Associates, Inc.},
 title = {{Perception Encoder}: The best visual embeddings are not at the output of the network},
 url = {https://proceedings.neurips.cc/paper_files/paper/2025/file/57bc0a850255e2041341bf74c7e2b9fa-Paper-Conference.pdf},
 volume = {38, Main Conference},
 year = {2025}
}

@inproceedings{ma2022x,
  author = {Ma, Yiwei and Xu, Guohai and Sun, Xiaoshuai and Yan, Ming and Zhang, Ji and Ji, Rongrong},
  title = {{X}-{CLIP}: End-to-End Multi-grained Contrastive Learning for Video-Text Retrieval},
  booktitle = {Proceedings of the 30th ACM International Conference on Multimedia},
  pages = {638--647},
  year = {2022},
  doi = {10.1145/3503161.3547910},
  url = {https://doi.org/10.1145/3503161.3547910},
}

@inproceedings{BLIP2,
  author = {Li, Junnan and Li, Dongxu and Savarese, Silvio and Hoi, Steven},
  title = {{BLIP}-2: Bootstrapping Language-Image Pre-training with Frozen Image Encoders and Large Language Models},
  booktitle = {Proceedings of the 40th International Conference on Machine Learning},
  volume = {202},
  pages = {19730--19742},
  year = {2023},
  series = {Proceedings of Machine Learning Research},
  month = {23--29 Jul},
  publisher = {PMLR},
  url = {https://proceedings.mlr.press/v202/li23q.html},
}

@inproceedings{BLIP,
  author = {Li, Junnan and Li, Dongxu and Xiong, Caiming and Hoi, Steven},
  title = {{BLIP}: Bootstrapping Language-Image Pre-training for Unified Vision-Language Understanding and Generation},
  booktitle = {Proceedings of the 39th International Conference on Machine Learning},
  volume = {162},
  pages = {12888--12900},
  year = {2022},
  series = {Proceedings of Machine Learning Research},
  month = {17--23 Jul},
  publisher = {PMLR},
  url = {https://proceedings.mlr.press/v162/li22n.html},
}

@inproceedings{CLIP,
  author = {Radford, Alec and Kim, Jong Wook and Hallacy, Chris and Ramesh, Aditya and Goh, Gabriel and Agarwal, Sandhini and Sastry, Girish and Askell, Amanda and Mishkin, Pamela and Clark, Jack and Krueger, Gretchen and Sutskever, Ilya},
  title = {Learning Transferable Visual Models From Natural Language Supervision},
  booktitle = {Proceedings of the 38th International Conference on Machine Learning},
  volume = {139},
  pages = {8748--8763},
  year = {2021},
  series = {Proceedings of Machine Learning Research},
  month = {18--24 Jul},
  publisher = {PMLR},
  url = {https://proceedings.mlr.press/v139/radford21a.html},
}

@inproceedings{devlin2019bert,
  author = {Devlin, Jacob and Chang, Ming-Wei and Lee, Kenton and Toutanova, Kristina},
  title = {{BERT}: Pre-training of Deep Bidirectional Transformers for Language Understanding},
  editor = {Burstein, Jill and Doran, Christy and Solorio, Thamar},
  booktitle = {Proceedings of the 2019 Conference of the North American Chapter of the Association for Computational Linguistics: Human Language Technologies, Volume 1 (Long and Short Papers)},
  month = jun,
  year = {2019},
  publisher = {Association for Computational Linguistics},
  pages = {4171--4186},
  doi = {10.18653/v1/N19-1423},
  url = {https://aclanthology.org/N19-1423/},
}

@article{luo2022clip4clip,
  author = {Luo, Huaishao and Ji, Lei and Zhong, Ming and Chen, Yang and Lei, Wen and Duan, Nan and Li, Tianrui},
  title = {{CLIP4Clip}: An empirical study of {CLIP} for end to end video clip retrieval and captioning},
  journal = {Neurocomputing},
  volume = {508},
  pages = {293--304},
  year = {2022},
  doi = {10.1016/j.neucom.2022.07.028},
  url = {https://doi.org/10.1016/j.neucom.2022.07.028},
  publisher = {Elsevier},
}

@inproceedings{khattab2020Colbert,
author = {Khattab, Omar and Zaharia, Matei}, 
title = {{ColBERT}: Efficient and Effective Passage Search via Contextualized Late Interaction over {BERT}}, 
url = {https://doi.org/10.1145/3397271.3401075}, 
doi = {10.1145/3397271.3401075},
year = {2020},   
booktitle = {Proceedings of the 43rd International ACM SIGIR Conference on Research and Development in Information Retrieval}, 
pages =  {39--48}, 
numpages = {10},
}

@inproceedings{yao2022filip,
  title={{FILIP}: Fine-grained Interactive Language-Image Pre-Training},
  author={Yao, Lewei and Huang, Runhui and Hou, Lu and Lu, Guansong and Niu, Minzhe and Xu, Hang and Liang, Xiaodan and Li, Zhenguo and Jiang, Xin and Xu, Chunjing},
  booktitle={International Conference on Learning Representations},
  year = {2022},
  url = {https://openreview.net/pdf?id=cpDhcsEDC2}
}

@inproceedings{Reddy_2025_CVPR,
  author = {Reddy, Arun and Martin, Alexander and Yang, Eugene and Yates, Andrew and Sanders, Kate and Murray, Kenton and Kriz, Reno and de Melo, Celso M. and Van Durme, Benjamin and Chellappa, Rama},
  title = {{Video-ColBERT}: Contextualized Late Interaction for Text-to-Video Retrieval},
  booktitle = {Proceedings of the IEEE/CVF Conference on Computer Vision and Pattern Recognition (CVPR)},
  pages = {19691--19701},
  year = {2025},
  month = {June},
  doi = {10.1109/cvpr52734.2025.01834},
  url = {https://doi.org/10.1109/cvpr52734.2025.01834},
}

@inproceedings{wang_etal_2024_videoclip,
  author = {Wang, Jiapeng and Wang, Chengyu and Huang, Kunzhe and Huang, Jun and Jin, Lianwen},
  editor = {Al-Onaizan, Yaser and Bansal, Mohit and Chen, Yun-Nung},
  title = {{V}ideo{CLIP}-{XL}: Advancing Long Description Understanding for Video {CLIP} Models},
  booktitle = {Proceedings of the 2024 Conference on Empirical Methods in Natural Language Processing},
  pages = {16061--16075},
  year = {2024},
  publisher = {Association for Computational Linguistics},
  address = {Miami, Florida, USA},
  month = nov,
  doi = {10.18653/v1/2024.emnlp-main.898},
  url = {https://aclanthology.org/2024.emnlp-main.898/},
}

@inproceedings{Xu_2016_CVPR,
  author = {Xu, Jun and Mei, Tao and Yao, Ting and Rui, Yong},
  title = {{MSR-VTT}: A Large Video Description Dataset for Bridging Video and Language},
  booktitle = {Proceedings of the IEEE Conference on Computer Vision and Pattern Recognition (CVPR)},
  year = {2016},
  month = {June},
  pages = {5288--5296},
  doi = {10.1109/cvpr.2016.571},
  url = {https://doi.org/10.1109/cvpr.2016.571},
}

@inproceedings{chen-dolan-2011-collecting,
  author = {Chen, David and Dolan, William},
  editor = {Lin, Dekang and Matsumoto, Yuji and Mihalcea, Rada},
  title = {Collecting Highly Parallel Data for Paraphrase Evaluation},
  booktitle = {Proceedings of the 49th Annual Meeting of the Association for Computational Linguistics: Human Language Technologies},
  month = jun,
  year = {2011},
  publisher = {Association for Computational Linguistics},
  pages = {190--200},
  address = {Portland, Oregon, USA},
  url = {https://aclanthology.org/P11-1020/},
}

@inproceedings{Krishna_2017_ICCV,
  author = {Krishna, Ranjay and Hata, Kenji and Ren, Frederic and Fei-Fei, Li and Niebles, Juan Carlos},
  title = {Dense-Captioning Events in Videos},
  booktitle = {Proceedings of the IEEE International Conference on Computer Vision (ICCV)},
  year = {2017},
  month = {October},
  pages = {706--715},
  doi = {10.1109/iccv.2017.83},
  url = {https://doi.org/10.1109/iccv.2017.83},
}

@inproceedings{Hendricks_2017_ICCV,
  author = {Hendricks, Lisa Anne and Wang, Oliver and Shechtman, Eli and Sivic, Josef and Darrell, Trevor and Russell, Bryan},
  title = {Localizing Moments in Video With Natural Language},
  booktitle = {Proceedings of the IEEE International Conference on Computer Vision (ICCV)},
  year = {2017},
  month = {October},
  pages = {5804--5813},
  doi = {10.1109/iccv.2017.618},
  url = {https://doi.org/10.1109/iccv.2017.618},
}

@inproceedings{taraday2026efficient,
title={Efficient Discriminative Joint Encoders for Large Scale Vision-Language Reranking},
author={Mitchell Keren Taraday and Shahaf Wagner and Chaim Baskin},
booktitle={The Fourteenth International Conference on Learning Representations},
year={2026},
url={https://openreview.net/pdf?id=UXtTBAyqVB}
}

@inproceedings{wang2020minilm_3f5ee243,
 author = {Wang, Wenhui and Wei, Furu and Dong, Li and Bao, Hangbo and Yang, Nan and Zhou, Ming},
 booktitle = {Advances in Neural Information Processing Systems},
 editor = {H. Larochelle and M. Ranzato and R. Hadsell and M.F. Balcan and H. Lin},
 pages = {5776--5788},
 publisher = {Curran Associates, Inc.},
 title = {{MiniLM}: Deep Self-Attention Distillation for Task-Agnostic Compression of Pre-Trained Transformers},
 url = {https://proceedings.neurips.cc/paper_files/paper/2020/file/3f5ee243547dee91fbd053c1c4a845aa-Paper.pdf},
 volume = {33},
 year = {2020}
}

@inproceedings{li2023unmasked,
  author = {Li, Kunchang and Wang, Yali and Li, Yizhuo and Wang, Yi and He, Yinan and Wang, Limin and Qiao, Yu},
  title = {Unmasked Teacher: Towards Training-Efficient Video Foundation Models},
  booktitle = {Proceedings of the IEEE/CVF International Conference on Computer Vision (ICCV)},
  month = {October},
  year = {2023},
  pages = {19948--19960},
  url = {https://openaccess.thecvf.com/content/ICCV2023/html/Li_Unmasked_Teacher_Towards_Training-Efficient_Video_Foundation_Models_ICCV_2023_paper.html}
}
\bibliographystyle{iclr2027_conference}

\clearpage
\appendix
\section{Additional Experimental Details}
\label{app:details}

This appendix documents the protocol behind the cached-reranking results and reports the secondary controls that complement the main experiments. \edit{Sections~\ref{app:algorithm}-\ref{app:controlled_comparisons} fix the deployment-relevant details that remain in the appendix (algorithm, architecture, training, cache quantization, compute, and code release); Sections~\ref{app:full_retrieval_results}-\ref{app:secondary_controls} provide the higher-rank cutoffs, the loss-stack ablation, the future-delta horizon sweep, and the reranker-backbone control.}

\edit{Recall values are percentages; absolute gains are percentage points.} \edit{Main-table $\pm$ entries report means and population standard deviations over three training seeds unless stated otherwise. The displayed SD has a minimum of 0.1 percentage points; smaller values are shown as 0.1.} \edit{Gains are differences between the reported recall values. Reported gain standard deviations describe variability across independent retrainings.} \edit{The default-budget $\deltaloss$ check uses four runs and reports the standard deviation of paired gains.}

\FloatBarrier
\subsection{\architecture{} Indexing and Reranking Algorithm}
\label{app:algorithm}

\edit{Algorithm~\ref{alg:vedje} specifies indexing and reranking.} \edit{Indexing encodes each video and stores its compressed frame tokens $Z(v)$.} \edit{Text-to-video retrieval encodes the query, retrieves candidates, fetches their tokens, and scores each query-video pair.} \edit{Video-to-text retrieval applies the same score $s_\theta(q,v)$ to candidate captions; a new video query first requires encoding and compression.}

\begin{algorithm}[H]
\caption{\edit{\architecture{}: indexing and text-to-video reranking.}}
\label{alg:vedje}
\begin{algorithmic}[1]
\ifshowrevisions\color{revisionorange}\fi
\Require frozen $g_\phi$; trained $C_\eta$, $W_{\mathrm{kv}}$, $Q_0$, joint encoder, $e_\rho$, $h_\psi$; first-stage retriever
\Statex \textbf{Offline indexing} (per video $v$):
\State sample $T$ frames and extract frame-indexed slices $X_t=[g_\phi(v)]_t$
\State $Z_t \gets C_\eta(Q_0;X_tW_{\mathrm{kv}})$ for $t=1,\ldots,T$ \Comment{Eq.~\ref{eq:frame_compressor}}
\State store ordered cache $Z(v) = [Z_1;\dots;Z_T]$ \Comment{Eq.~\ref{eq:ordered_cache}}
\Statex \textbf{Online reranking} (per query $q$):
\State $\mathcal{C}(q),\,\rho(q,\cdot) \gets$ first-stage text encoding and search
\State fetch cached tokens for $v\in\mathcal{C}(q)$
\For{each candidate $v\in\mathcal{C}(q)$ (batched in implementation)}
    \State $c(q,v) \gets$ joint encoder over text and cache embeddings with sequence positions
    \State $s_\theta(q,v) \gets h_\psi\!\big(c(q,v) + e_\rho(\rho(q,v))\big)$ \Comment{Eq.~\ref{eq:residual_score}}
\EndFor
\State \Return $\mathcal{C}(q)$ sorted by $s_\theta$ \Comment{no $g_\phi$ on the query path}
\end{algorithmic}
\end{algorithm}

\FloatBarrier
\subsection{Architecture Details}
\label{app:arch_details}

\edit{This subsection specifies the components in Section~\ref{sec:method}; symbol use matches the main text.} \edit{Table~\ref{tab:backbone_sources} identifies the feature and contrastive-target sources for the main variants and the CLIP control. The dimensions and checkpoint details below describe the default VideoPrism configuration.}

\begin{table}[htbp]
\centering
\caption{\edit{Backbone sources for the main variants and the CLIP ViT-B/16 control. Each row identifies the first-stage model, the source of cached visual features, and the source of frozen video embeddings for VTC. ZS and FT describe first-stage training.}}
\label{tab:backbone_sources}
\begin{editblock}
\begingroup
\setlength{\tabcolsep}{4pt}
\renewcommand{\arraystretch}{1.08}
\begin{tabularx}{\linewidth}{@{}XXX@{}}
\toprule
First-stage model & Cache features & VTC video targets \\
\midrule
VideoPrism-LvT-B & VideoPrism-Base & VideoPrism-LvT-B \\
VideoCLIP-XL (ZS) & VideoCLIP-XL & VideoCLIP-XL \\
VideoCLIP-XL (FT) & VideoCLIP-XL & VideoCLIP-XL \\
PE-Core-B & PE-Core-B & PE-Core-B \\
CLIP ViT-B/16 (ZS) & CLIP ViT-B/16 & CLIP ViT-B/16 \\
\bottomrule
\end{tabularx}
\endgroup
\end{editblock}
\end{table}

\paragraph{Visual backbone $g_\phi$.} VideoPrism-Base~\citep{zhao2024videoprism}, used at $288{\times}288$ resolution with 16 frames per video. The HuggingFace checkpoint is \texttt{MHRDYN7/videoprism-base-f16r288}. The backbone is frozen at every stage: gradients do not flow into its weights, and it runs once per video at indexing time. \edit{Its feature tensor is partitioned into $T$ frame-indexed slices, each with $P{=}256$ spatial positions and $d_v{=}768$ channels.} \edit{The VideoPrism architecture includes temporal attention, so a frame-indexed slice need not contain information from that frame alone.}

\paragraph{\edit{Frame-indexed compressor.}} A 2-layer transformer-decoder stack operates in the joint-encoder dimension $d_\ell{=}384$. \edit{The stack $C_\eta$ in Eq.~\ref{eq:frame_compressor} uses learned queries $Q_0\in\mathbb{R}^{M\times d_\ell}$ and projected frame features $U_t=X_tW_{\mathrm{kv}}$ as its memory.} \edit{The shared projection $W_{\mathrm{kv}}\in\mathbb{R}^{d_v\times d_\ell}$ is \texttt{kv\_proj}.} The two decoder layers use 8 attention heads, FFN width $4d_\ell{=}1536$, GELU activation, and dropout 0.1. The output of the second layer is the per-frame cache $Z_t\in\mathbb{R}^{M\times d_\ell}$. Concatenating frames in temporal order yields $Z(v) {=} [Z_1;\dots;Z_T] \in \mathbb{R}^{TM\times d_\ell}$. The default $TM{=}64$ uses $T{=}16$, $M{=}4$. Compact configurations reduce $M$ while keeping $T{=}16$ (e.g., $TM{=}16$ uses $M{=}1$).

\paragraph{Joint reranker (online).} \texttt{microsoft/MiniLM-L12-H384-uncased}~\citep{wang2020minilm_3f5ee243}: 12 transformer blocks, hidden size $d_\ell{=}384$, 12 attention heads, $\sim$33M parameters, with the standard BERT~\citep{devlin2019bert} uncased WordPiece tokenizer (max text length 64 tokens, including \texttt{[CLS]} and \texttt{[SEP]}). Cached video tokens $Z(v)$ are concatenated with the text input embeddings as additional input positions. \edit{The reranker's learned sequence-position embeddings cover the concatenated text and video positions; the cache itself is stored before this positional addition.}

\paragraph{Score-encoder $e_\rho$.} The scalar stage-1 score $\rho(q,v)$ is mapped to $\mathbb{R}^{d_\ell}$ by a 2-layer MLP $\mathrm{Linear}(1, 64)\to\mathrm{GELU}\to\mathrm{Linear}(64, d_\ell)$.

\edit{\paragraph{Score head.} The head $h_\psi$ in Eq.~\ref{eq:residual_score} is $\mathrm{Linear}(d_\ell,d_\ell)\to\mathrm{GELU}\to\mathrm{Linear}(d_\ell,1)$, with approximately 0.15M parameters.} \edit{The score encoder and score head together add fewer than 0.2M parameters beyond the approximately 33M joint encoder.}

\paragraph{Future-delta predictor $F_\omega$ (training only).} A 2-layer transformer-decoder stack with 8 attention heads, FFN width $4d_\ell{=}1536$, GELU activation, dropout 0.1, and LayerNorm on the output. The default horizon set is the single horizon $\mathcal{H}{=}\{3\}$ (Section~\ref{app:horizon_sweep} reports $h\in\{2,3,4\}$). Queries are \emph{fixed} sinusoidal vectors of dimension $d_\ell$ indexed by horizon and patch position $(h, p)$ with $h\in\mathcal{H}$ and $p\in\{0,\dots,P{-}1\}$ ($P{=}256$); the first half of $d_\ell$ encodes $h$ and the second half encodes $p$ via standard sinusoidal positional encoding. The decoder cross-attends to a single frame's compressed tokens $Z_t\in\mathbb{R}^{M\times d_\ell}$, and a final linear layer projects each output to $\mathbb{R}^{d_v}{=}\mathbb{R}^{768}$ to produce $\widehat{\Delta}_{t}\in\mathbb{R}^{|\mathcal{H}|\times P\times d_v}$. \edit{The predictor is discarded at deployment, so its parameter count does not affect the 33M added reranker cost.}

\edit{\paragraph{Cache layout and storage.} The cached tensor has shape $TM\times d_\ell$.} \edit{Its BF16 payload is $TMd_\ell\cdot2$ bytes: 49,152 bytes (48\,KiB) for 64 tokens and 12,288 bytes (12\,KiB) for 16 tokens at $d_\ell=384$.} \edit{A KiB contains 1,024 bytes; kB and larger SI units are decimal.} \edit{These payloads exclude file-container and index metadata.}

\FloatBarrier
\subsection{Training Protocol}
\label{app:training_protocol}

Training is a single stage that jointly optimizes the four objectives in Eq.~\ref{eq:total_loss}: the matching loss $\mathcal{L}_{\mathrm{vtm}}$, the contrastive loss $\mathcal{L}_{\mathrm{vtc}}$, the masked-language loss $\mathcal{L}_{\mathrm{mlm}}$, and the future-delta auxiliary $\deltaloss{}$.

\edit{\paragraph{Trainable and frozen parameters.} The visual backbone $g_\phi$ remains frozen.} \edit{Training jointly updates the compressor, joint encoder, score encoder $e_\rho$, score head $h_\psi$, VTC text projection, MLM head, and future-delta predictor.} \edit{The final cache is written with the trained compressor.} \edit{At deployment the auxiliary heads are discarded, and joint scoring uses the cached tokens after candidate search.} \edit{An update to the visual backbone or compressor requires rewriting the cache; an update confined to the online scorer can reuse it.}

\paragraph{Optimization.}
\begin{itemize}
    \item Optimizer: AdamW with default $\beta{=}(0.9, 0.999)$ and weight decay $0.02$.
    \item Schedule: 400 warmup steps from $10^{-6}$ to $3\times 10^{-4}$, then per-epoch step decay with multiplicative factor $0.9$.
    \item Batch size: 64 query-video pairs per optimizer step.
    \item Epochs: 4.
\end{itemize}

\paragraph{Loss definitions used in training.}
\begin{itemize}
    \item \emph{VTM} ($\mathcal{L}_{\mathrm{vtm}}$): softmax cross-entropy over the candidate scores $s_\theta(q,v_i)$ at temperature $\tau_{\mathrm{vtm}}{=}1.0$, with the positive as the target.
    \item \emph{VTC} ($\mathcal{L}_{\mathrm{vtc}}$): symmetric InfoNCE between the text CLS embedding and frozen video embeddings, at logit scale 20.0. \edit{In the default VideoPrism configuration, text features are linearly projected to $d_{\mathrm{clip}}{=}768$ and both modalities are L2-normalized; video targets come from VideoPrism-LvT-B~\citep{zhao2024videoprism}. Other variants use video targets from their corresponding backbone, as specified in Table~\ref{tab:backbone_sources}.}
    \item \emph{MLM} ($\mathcal{L}_{\mathrm{mlm}}$): standard masked-language modeling on the text tokens, computed jointly with the cached video tokens in context (i.e., the encoder reads both modalities).
    \item \emph{\deltaloss{}}: squared $\ell_2$ regression as in Eq.~\ref{eq:delta_loss}, mean-squared error between the predicted patch-level delta $\widehat{\Delta}_{t,h}$ and the frozen-backbone target $\Delta_{t,h}{=}X_{t+h}{-}X_t$, averaged over patches and the visual feature dimension and then over the set of valid future pairs $\Omega = \{(t,h) : 1\le t,\, t+h\le T,\,h\in\mathcal{H}\}$. The default horizon set is the single horizon $\mathcal{H}{=}\{3\}$; Section~\ref{app:horizon_sweep} reports $h\in\{2,3,4\}$.
\end{itemize}

\FloatBarrier
\subsection{Cache Quantization}
\label{app:quantization_details}

\edit{The default cache uses BF16.} \edit{The FP8-E4M3 and FP4-E2M1 controls in Table~\ref{tab:app_token_precision_details} quantize the stored cache and dequantize it to FP32 before joint encoding.} \edit{Reranker and score-encoder weights retain their training precision.} \edit{Storage counts the nominal packed payload: 0.5, 1, or 2 bytes per element for FP4, FP8, or BF16, excluding metadata.} \edit{The controls measure retrieval accuracy after cache quantization, with floating-point inference in every condition.}

\FloatBarrier
\subsection{Compute and Code Release}
\label{app:compute_code_release}

\edit{\paragraph{Hardware.} Training and indexing use one 48\,GB NVIDIA L40S GPU.} \edit{A default four-epoch run with batch size 64 takes approximately 16 to 24 hours.} \edit{The compute estimate of 850 to 900 GPU-hours covers the initial configurations, seed repeats, and appendix ablations.} \edit{Preliminary and failed runs are excluded, and compute for the additional controlled comparisons is not fully accounted for in this estimate.}

\edit{\paragraph{Code release.} \githubcodelink}
\edit{\paragraph{License and asset notes.} The official \href{https://huggingface.co/google/videoprism-base-f16r288}{VideoPrism-Base} and \href{https://huggingface.co/google/videoprism-lvt-base-f16r288}{VideoPrism-LvT-B} model cards~\citep{zhao2024videoprism} declare Apache License 2.0.} \edit{The experiments use the public HuggingFace mirrors \path{MHRDYN7/videoprism-base-f16r288} and \path{MHRDYN7/videoprism-lvt-base-f16r288}; their pages carry no additional license declaration.} \edit{The MiniLM~\citep{wang2020minilm_3f5ee243} checkpoint \path{microsoft/MiniLM-L12-H384-uncased} is distributed under the MIT license.} MSR-VTT, MSVD, DiDeMo, and ActivityNet are used under their original release terms.

\FloatBarrier
\subsection{\edit{\texorpdfstring{End-to-End Latency and Cache I/O}{End-to-End Latency and Cache I/O}}}
\label{app:latency_io_protocol}

\edit{\paragraph{Timing protocol.} We time VP-backed MSR-VTT T2V retrieval on one NVIDIA L40S with one query per batch and all its candidates processed together.} \edit{Measurements are medians of 200 runs after 30 warmups.} \edit{The approximately 4.3\,ms stage-1 reference includes query-text encoding and 0.03 to 0.04\,ms candidate search.} \edit{End-to-end timings additionally include prior processing, cache fetching and host-to-device transfer, and joint scoring.} \edit{With caches already on device, the FP16 reranker takes 3.03/1.73\,ms for 64/16 tokens at $N_{\mathrm{c}}{=}100$; PyTorch CUDA peak resident memory is 0.99/0.78\,GB for this component.}

\begin{table}[htbp]

    \centering
    \caption{\edit{VP-backed MSR-VTT T2V latency for 64-token caches. Values are milliseconds, measured end to end under host-RAM and batched NVMe reads; warm and cold denote the reported cache-residency conditions.}}
    \label{tab:app_latency_io}
    \begingroup
    \normalsize
    \setlength{\tabcolsep}{8pt}
    \renewcommand{\arraystretch}{1.12}
    \arrayrulecolor{gray!70}

\begin{editblock}
\begin{tabularx}{\linewidth}{>{\raggedright\arraybackslash}X r r}
        \toprule
        \textbf{Cache path} & \textbf{$N_{\mathrm{c}}{=}20$} & \textbf{$N_{\mathrm{c}}{=}100$} \\
        \midrule
        Host RAM & 5.79 & 7.83 \\
        NVMe, warm & 6.3 & 7.8 \\
        NVMe, cold & 7.4 & 11.1 \\
        \bottomrule
    \end{tabularx}
\end{editblock}

    \arrayrulecolor{black}
    \endgroup

\end{table}

\edit{At $N_{\mathrm{c}}{=}20$, host-cache fetching and copying accounts for 2.4\%/1.2\% of end-to-end query time for 64/16 tokens; the share remains at most 5.5\% through 100 candidates.} \edit{Warm NVMe reads approach host-RAM latency, while cold reads are slower.} \edit{\paragraph{Payload extrapolation.} Multiplying the per-video payload by corpus size gives 12.3/49.2\,GB for one million videos and 0.123/0.492\,TB for ten million at 16/64 BF16 tokens.} \edit{At 16 tokens, nominal packed FP4 payload is 3.07\,GB per million videos.} \edit{These calculations count cached tensor bytes, excluding container metadata and the first-stage index; they are storage extrapolations.}

\edit{\paragraph{Visual representation storage.} At two bytes per element, storing 256 patch vectors and one frame vector for each of 16 sampled frames requires 6.02\,MiB per video. \architecture{} stores 64 vectors of dimension 384, reducing this payload to 48\,KiB, a factor of 128.5. The 16-token configuration uses 12\,KiB.}

\begin{table}[htbp]
\centering
\caption{\edit{\textbf{Per-video visual storage at two bytes per element.} The full-feature reference contains $16\times257\times768$ elements; \architecture{} contains $64\times384$ or $16\times384$. Values count tensor payload. The first-stage search index and file metadata are accounted for separately.}}
\label{tab:app_visual_payload}
\begin{editblock}
\begin{tabularx}{\linewidth}{X r}
\toprule
Representation & Visual payload per video \\
\midrule
Uncompressed VideoPrism frame-and-patch representation & 6.02\,MiB \\
\architecture{}, 64 tokens & 48\,KiB \\
\architecture{}, 16 tokens & 12\,KiB \\
\bottomrule
\end{tabularx}
\end{editblock}
\end{table}

\edit{\paragraph{Measurement scope.} Component timings and memory cover the reranker; end-to-end timings cover the complete query under the stated hardware, batching, and cache-residency conditions.}

\FloatBarrier
\subsection{\edit{\texorpdfstring{Controlled Comparison Protocols}{Controlled Comparison Protocols}}}
\label{app:controlled_comparisons}

\phantomsection\label{app:single_frame_reference}
\edit{\paragraph{Single-frame EDJE-style reference.} The EDJE-style reference~\citep{taraday2026efficient} compresses the middle sampled frame into a 64-token cache and evaluates a joint scorer without the first-stage score input. It reaches $29.6\pm0.4$ T2V and $31.4\pm0.6$ V2T R@1 on the same VideoPrism candidates (mean $\pm$ standard deviation over three training seeds). This single-frame reference is reported separately from the matched multiframe cache-structure comparison in Table~\ref{tab:cache_structure}.}

\phantomsection\label{app:four_run_auxiliary}
\edit{\paragraph{Four-run auxiliary check.} A separate check on MSR-VTT at the default 64-token budget compares training with and without \deltaloss{} over four runs. Gains are positive in every run, with mean T2V/V2T gains of $+0.60/+1.00$ R@1 points and standard deviations of paired gains of $0.13/0.11$. This check is separate from the three-seed cache-budget comparison in Table~\ref{tab:token_budget_storage}.}

\edit{\paragraph{Same-cache scoring.} Table~\ref{tab:main_same_cache_scoring} reports the comparison. The two rerankers use the same VideoPrism features, 64-token cache, first stage, and 20-candidate pool.} \edit{Raw VideoPrism supplies the stage-1 reference.} \edit{The trained MaxSim head follows the same training data and schedule as \architecture{}.}

\edit{\paragraph{Released EERCF candidate generator.} We use EERCF's released CLIP ViT-B/32 checkpoint~\citep{tian2024towards} to generate 50 candidates per query on MSR-VTT.} \edit{Our evaluation reproduces its coarse T2V R@1 of 42.8; \architecture{} reranks these candidates to 49.1.} \edit{Table~\ref{tab:app_eercf_native_control} also includes EERCF's published 47.8 R@1 under its own reranking protocol.}

\begin{table}[htbp]

    \centering
    \caption{\edit{MSR-VTT T2V using the released EERCF CLIP ViT-B/32 candidate generator ($N_{\mathrm{c}}{=}50$). Stage 1 and VEDJE are evaluated here; the EERCF score is the published reference.}}
    \label{tab:app_eercf_native_control}
    \begingroup
    \normalsize
    \setlength{\tabcolsep}{10pt}
    \renewcommand{\arraystretch}{1.12}
    \arrayrulecolor{gray!70}

\begin{editblock}
\begin{tabularx}{\linewidth}{>{\raggedright\arraybackslash}X r}
        \toprule
        \textbf{Method} & \textbf{\shortstack{T2V\\R@1}} \\
        \midrule
        Released coarse checkpoint (reproduced) & 42.8 \\
        EERCF published & 47.8 \\
        \oursrow
        \textbf{\architecture{} on released-checkpoint candidates} & \textbf{49.1} \\
        \bottomrule
    \end{tabularx}
\end{editblock}

    \arrayrulecolor{black}
    \endgroup

\end{table}

\edit{\paragraph{Matched-backbone and stage-1 training controls.} Table~\ref{tab:controlled_comparisons} uses CLIP ViT-B/16.} \edit{\architecture{} freezes the public checkpoint and uses zero-shot stage 1; the published comparators fine-tune that backbone.} \edit{The cross-system CLIP4Clip rows use B/32 and are reported separately.} \edit{The fine-tuned control uses our VideoCLIP-XL stage 1 on MSR-VTT: T2V R@1/R@5/R@10 is 56.2/80.2/87.4 and V2T is 55.1/81.8/88.5, against the original paper's 57.0 T2V R@1~\citep{wang_etal_2024_videoclip}.} \edit{We regenerate candidates and priors and retrain the reranker using VideoCLIP-XL cache features.} \edit{The reranked T2V/V2T R@1 is 59.8/58.8; evaluation of this condition is reported at R@1.}

\begin{table}[htbp]
    \centering
    \caption{\edit{Stage-1 prior controls on MSR-VTT 1ka with fixed candidates. Prior-only and joint-only models are trained independently using their respective inputs. The full model combines the prior with joint text-video features.}}
    \label{tab:stage1_prior_controls}
    \begingroup
    \normalsize
    \setlength{\tabcolsep}{5pt}
    \renewcommand{\arraystretch}{1.10}
    \arrayrulecolor{gray!70}

\begin{editblock}
\begin{tabularx}{\linewidth}{>{\raggedright\arraybackslash}X c c r r}
        \toprule
        \textbf{Variant} & \textbf{\shortstack{Joint\\tokens?}} & \textbf{\shortstack{Learned\\prior?}} & \textbf{\shortstack{T2V\\R@1}} & \textbf{\shortstack{V2T\\R@1}} \\
        \midrule
        Stage-1 baseline & \xmark & \xmark & 50.1 & 49.8 \\
        Prior stream only & \xmark & \cmark & 50.1 & 49.8 \\
        Joint stream only & \cmark & \xmark & 43.8 & 41.6 \\
        \oursrow
        \textbf{Full \architecture{}} & \cmark & \cmark & \textbf{54.6} & \textbf{54.6} \\
        \bottomrule
    \end{tabularx}
\end{editblock}

    \arrayrulecolor{black}
    \endgroup
\end{table}

\FloatBarrier
\subsection{Full Retrieval Results}
\label{app:full_retrieval_results}

\edit{Tables~\ref{tab:app_full_accuracy_t2v} and~\ref{tab:app_full_accuracy_v2t} report Recall@$K$ for text-to-video and video-to-text retrieval separately, in percent.} The variant convention follows the main text: \architecture{} variants are written \vedjevariant{S1}{N}, with the subscript naming the \edit{retrieval and cache backbone} and $N$ naming cached tokens (omitted in print when $N{=}64$). Table~\ref{tab:app_matched_reranking} then isolates the matched first-stage comparison and reports the same gain explicitly.

\begin{table}[htbp]
\centering
\caption{\edit{\textbf{Cross-dataset retrieval before and after VEDJE.} R@1 is in percent; gains are in percentage points. Both panels use 64-token caches. Full retrieval results appear in Table~\ref{tab:app_matched_reranking}.}}
\label{tab:cross_dataset_r1}
\begin{editblock}
\begingroup
\setlength{\tabcolsep}{4pt}
\renewcommand{\arraystretch}{1.0}
\begin{tabularx}{\linewidth}{@{}l >{\raggedleft\arraybackslash}X >{\columncolor{ourmethodrow}\raggedleft\arraybackslash}X >{\raggedleft\arraybackslash}X >{\raggedleft\arraybackslash}X >{\columncolor{ourmethodrow}\raggedleft\arraybackslash}X >{\raggedleft\arraybackslash}X@{}}
\toprule
& \multicolumn{3}{c}{\textbf{Text-to-video}} & \multicolumn{3}{c@{}}{\textbf{Video-to-text}} \\
\cmidrule(lr){2-4}\cmidrule(l){5-7}
Dataset & Stage 1 & VEDJE & Gain & Stage 1 & VEDJE & Gain \\
\midrule
\multicolumn{7}{@{}l}{\textbf{VideoPrism first stage}} \\
MSR-VTT & 50.1 & \textbf{54.6} & +4.5 & 49.8 & \textbf{54.6} & +4.8 \\
MSVD & 55.6 & \textbf{56.7} & +1.1 & 83.4 & \textbf{83.7} & +0.3 \\
DiDeMo & 47.1 & \textbf{48.2} & +1.1 & 47.3 & \textbf{47.9} & +0.6 \\
ActivityNet & 48.8 & \textbf{50.6} & +1.8 & 47.9 & \textbf{48.3} & +0.4 \\
\midrule
\multicolumn{7}{@{}l}{\textbf{VideoCLIP-XL first stage (zero-shot)}} \\
MSR-VTT & 50.1 & \textbf{56.5} & +6.4 & 49.9 & \textbf{57.1} & +7.2 \\
MSVD & 51.9 & \textbf{56.8} & +4.9 & 76.7 & \textbf{76.9} & +0.2 \\
DiDeMo & 47.7 & \textbf{56.6} & +8.9 & 47.9 & \textbf{54.8} & +6.9 \\
ActivityNet & 46.4 & \textbf{51.6} & +5.2 & 48.1 & \textbf{49.9} & +1.8 \\
\bottomrule
\end{tabularx}
\endgroup
\end{editblock}
\end{table}

\FloatBarrier

\edit{VEDJE results are ours; VP, VCLIP, and PE identify the first-stage retrievers VideoPrism~\citep{zhao2024videoprism}, VideoCLIP-XL~\citep{wang_etal_2024_videoclip}, and Perception Encoder~\citep{bolya2025perceptionencoder}.} \edit{VCLIP uses zero-shot stage 1 unless marked fine-tuned.} \edit{Published CLIP4Clip cross-dataset rows use ViT-B/32; the separate matched-backbone control uses ViT-B/16.} \edit{LamRA values are the VideoChat-Flash reproduction reported by \citet{lee2025captioning}.} \edit{Dashes denote unreported values.}

\begin{table}[htbp]
    \centering
    \caption{\edit{Full text-to-video retrieval accuracy across MSR-VTT, MSVD, DiDeMo, and ActivityNet. Recall@$K$ is in percent. Variant and reporting conventions are specified above.}}
    \label{tab:app_full_accuracy_t2v}
    \begingroup
    \scriptsize
    \setlength{\tabcolsep}{2.8pt}
    \renewcommand{\arraystretch}{1.08}
    \arrayrulecolor{gray!70}
    \begin{editblock}
    \begin{adjustbox}{max width=\linewidth}
    \begin{tabular}{@{}l *{12}{c}@{}}
        \toprule
        & \multicolumn{3}{c}{\textbf{MSR-VTT}} & \multicolumn{3}{c}{\textbf{MSVD}} & \multicolumn{3}{c}{\textbf{DiDeMo}} & \multicolumn{3}{c}{\textbf{ActivityNet}} \\
        \cmidrule(lr){2-4}\cmidrule(lr){5-7}\cmidrule(lr){8-10}\cmidrule(l){11-13}
        \textbf{Method} & \textbf{R@1} & \textbf{R@5} & \textbf{R@10} & \textbf{R@1} & \textbf{R@5} & \textbf{R@10} & \textbf{R@1} & \textbf{R@5} & \textbf{R@10} & \textbf{R@1} & \textbf{R@5} & \textbf{R@10} \\
        \midrule
        \multicolumn{13}{@{}l}{\emph{Reported retrieval baselines}} \\
        \addlinespace[1pt]
        \methodcell{CLIP4Clip (B/32)}{luo2022clip4clip} & 44.5 & 71.4 & 81.6 & 45.2 & 75.5 & -- & 42.8 & 68.5 & 79.2 & 40.5 & 72.4 & -- \\
        \methodcell{X-CLIP (B/16)}{ma2022x} & 49.3 & 75.8 & 84.8 & 50.4 & 80.6 & -- & 47.8 & 79.3 & -- & 46.2 & 75.5 & -- \\
        \methodcell{TeachCLIP}{tian2023teachclip} & 45.2 & 72.3 & 82.3 & 47.4 & 77.3 & 85.5 & -- & -- & -- & 42.2 & 72.7 & 85.2 \\
        \methodcell{Video-ColBERT}{Reddy_2025_CVPR} & 51.5 & 76.3 & 85.5 & 55.2 & 82.9 & 89.4 & 51.7 & 76.1 & 84.8 & 45.8 & 76.3 & 86.7 \\
        \midrule
        \multicolumn{13}{@{}l}{\emph{Online rerankers and MLLM scorers}} \\
        \addlinespace[1pt]
        \methodcell{CrossTVR-Large}{DAI2025111099} & 54.0 & 77.5 & 85.3 & 55.0 & 81.9 & -- & 55.0 & 77.6 & -- & -- & -- & -- \\
        \methodcell{LamRA}{liu2025lamra} & 59.7 & 81.4 & 87.2 & 59.0 & -- & -- & 83.5 & 94.8 & 96.2 & 76.0 & 92.8 & 96.3 \\
        \methodcell{CaRe-DPO}{lee2025captioning} & 64.1 & 83.8 & 88.8 & 59.8 & -- & -- & 85.1 & 95.0 & 96.2 & 79.2 & 93.6 & 96.5 \\
        \midrule
        \multicolumn{13}{@{}l}{\emph{Cached joint reranking}} \\
        \addlinespace[1pt]
        \oursrow
        \vedjevariantbf{VP}{64} & 54.6 & 76.9 & 85.4 & 56.7 & 83.7 & 89.8 & 48.2 & 72.9 & 80.3 & 50.6 & 77.8 & 87.7 \\
        \oursrow
        \vedjevariantbf{VCLIP}{64} & 56.5 & 81.8 & 88.0 & 56.8 & 83.0 & 90.1 & 56.6 & 79.8 & 86.8 & 51.6 & 79.9 & 89.3 \\
        \oursrow
        \vedjevariantbf{PE}{64} & 53.9 & 77.3 & 85.0 & -- & -- & -- & -- & -- & -- & -- & -- & -- \\
        \bottomrule
    \end{tabular}
    \end{adjustbox}
    \end{editblock}
    \arrayrulecolor{black}
    \endgroup
\end{table}

\begin{table}[htbp]
    \centering
    \caption{\edit{Full video-to-text retrieval accuracy across MSR-VTT, MSVD, DiDeMo, and ActivityNet. Recall@$K$ is in percent; variant and reporting conventions follow Table~\ref{tab:app_full_accuracy_t2v}.}}
    \label{tab:app_full_accuracy_v2t}
    \begingroup
    \scriptsize
    \setlength{\tabcolsep}{2.8pt}
    \renewcommand{\arraystretch}{1.08}
    \arrayrulecolor{gray!70}
    \begin{editblock}
    \begin{adjustbox}{max width=\linewidth}
    \begin{tabular}{@{}l *{12}{c}@{}}
        \toprule
        & \multicolumn{3}{c}{\textbf{MSR-VTT}} & \multicolumn{3}{c}{\textbf{MSVD}} & \multicolumn{3}{c}{\textbf{DiDeMo}} & \multicolumn{3}{c}{\textbf{ActivityNet}} \\
        \cmidrule(lr){2-4}\cmidrule(lr){5-7}\cmidrule(lr){8-10}\cmidrule(l){11-13}
        \textbf{Method} & \textbf{R@1} & \textbf{R@5} & \textbf{R@10} & \textbf{R@1} & \textbf{R@5} & \textbf{R@10} & \textbf{R@1} & \textbf{R@5} & \textbf{R@10} & \textbf{R@1} & \textbf{R@5} & \textbf{R@10} \\
        \midrule
        \multicolumn{13}{@{}l}{\emph{Reported retrieval baselines}} \\
        \addlinespace[1pt]
        \methodcell{CLIP4Clip (B/32)}{luo2022clip4clip} & 43.1 & 70.5 & 81.2 & 62.0 & 87.3 & -- & 42.5 & 70.6 & 80.2 & 42.6 & 73.4 & -- \\
        \methodcell{X-CLIP (B/16)}{ma2022x} & 48.9 & 76.8 & 84.5 & 66.8 & 90.4 & -- & 47.8 & 76.8 & -- & 46.4 & 75.9 & -- \\
        \methodcell{TeachCLIP}{tian2023teachclip} & -- & -- & -- & -- & -- & -- & -- & -- & -- & -- & -- & -- \\
        \methodcell{Video-ColBERT}{Reddy_2025_CVPR} & -- & -- & -- & -- & -- & -- & -- & -- & -- & -- & -- & -- \\
        \midrule
        \multicolumn{13}{@{}l}{\emph{Online rerankers and MLLM scorers}} \\
        \addlinespace[1pt]
        \methodcell{CrossTVR-Large}{DAI2025111099} & 51.3 & 78.3 & 85.4 & 74.3 & 95.5 & -- & 51.3 & 76.7 & -- & -- & -- & -- \\
        \methodcell{LamRA}{liu2025lamra} & 60.7 & 82.3 & 89.0 & 85.5 & -- & -- & 79.4 & 94.8 & 96.6 & 68.7 & 90.1 & 95.3 \\
        \methodcell{CaRe-DPO}{lee2025captioning} & 63.8 & 83.0 & 87.3 & 85.9 & -- & -- & 82.5 & 95.2 & 96.3 & 74.4 & 92.4 & 96.3 \\
        \midrule
        \multicolumn{13}{@{}l}{\emph{Cached joint reranking}} \\
        \addlinespace[1pt]
        \oursrow
        \vedjevariantbf{VP}{64} & 54.6 & 79.3 & 86.7 & 83.7 & 96.0 & 98.7 & 47.9 & 73.5 & 81.5 & 48.3 & 76.8 & 86.6 \\
        \oursrow
        \vedjevariantbf{VCLIP}{64} & 57.1 & 81.6 & 88.7 & 76.9 & 93.8 & 96.4 & 54.8 & 80.8 & 86.4 & 49.9 & 79.1 & 89.5 \\
        \oursrow
        \vedjevariantbf{PE}{64} & 53.9 & 78.2 & 85.9 & -- & -- & -- & -- & -- & -- & -- & -- & -- \\
        \bottomrule
    \end{tabular}
    \end{adjustbox}
    \end{editblock}
    \arrayrulecolor{black}
    \endgroup
\end{table}

\begin{table}[htbp]
    \centering
    \caption{\edit{Matched first-stage and cached-reranker retrieval. Recall@$K$ is in percent; gains are absolute R@1 changes in percentage points. Each panel fixes the candidate generator (named by the \architecture{} subscript) and reranker. On MSR-VTT, three-seed T2V R@1 is $54.6\pm0.1$ for VP and $56.5\pm0.2$ for VCLIP (mean $\pm$ standard deviation). VideoPrism V2T gains on MSVD, DiDeMo, and ActivityNet were evaluated across independent retrainings. FT denotes a fine-tuned first stage.}}
    \label{tab:app_matched_reranking}
    \begingroup
    \normalsize
    \setlength{\tabcolsep}{3pt}
    \renewcommand{\arraystretch}{1.15}
    \arrayrulecolor{gray!70}
    \begin{editblock}
    \begin{tabularx}{\linewidth}{>{\raggedright\arraybackslash}X c r r >{\columncolor{ourmethodrow}}c r r}
        \toprule
        \textbf{Dataset} & \textbf{Dir.} & \textbf{\shortstack{Stage-1\\R@1}} & \textbf{\shortstack{Reranked\\R@1}} & \textbf{Gain} & \textbf{\shortstack{Reranked\\R@5}} & \textbf{\shortstack{Reranked\\R@10}} \\
        \midrule
        \multicolumn{7}{@{}p{\linewidth}@{}}{\emph{VideoPrism-LvT-B~\citep{zhao2024videoprism} stage 1 $\rightarrow$ \vedjevariant{VP}{64}}} \\
        \addlinespace[1pt]
        MSR-VTT & T2V & 50.1 & 54.6 & +4.5 & 76.9 & 85.4 \\
        MSR-VTT & V2T & 49.8 & 54.6 & +4.8 & 79.3 & 86.7 \\
        \addlinespace[1pt]
        MSVD & T2V & 55.6 & 56.7 & +1.1 & 83.7 & 89.8 \\
        MSVD & V2T & 83.4 & 83.7 & +0.3 & 96.0 & 98.7 \\
        \addlinespace[1pt]
        DiDeMo & T2V & 47.1 & 48.2 & +1.1 & 72.9 & 80.3 \\
        DiDeMo & V2T & 47.3 & 47.9 & +0.6 & 73.5 & 81.5 \\
        \addlinespace[1pt]
        ActivityNet & T2V & 48.8 & 50.6 & +1.8 & 77.8 & 87.7 \\
        ActivityNet & V2T & 47.9 & 48.3 & +0.4 & 76.8 & 86.6 \\
        \midrule
        \multicolumn{7}{@{}p{\linewidth}@{}}{\emph{PE-Core-B~\citep{bolya2025perceptionencoder} stage 1 $\rightarrow$ \vedjevariant{PE}{64}}} \\
        \addlinespace[1pt]
        MSR-VTT & T2V & 47.6 & 53.9 & +6.3 & 77.3 & 85.0 \\
        MSR-VTT & V2T & 47.3 & 53.9 & +6.6 & 78.2 & 85.9 \\
        \midrule
        \multicolumn{7}{@{}p{\linewidth}@{}}{\emph{VideoCLIP-XL~\citep{wang_etal_2024_videoclip} stage 1 (zero-shot unless noted) $\rightarrow$ \vedjevariant{VCLIP}{64}}} \\
        \addlinespace[1pt]
        MSR-VTT & T2V & 50.1 & 56.5 & +6.4 & 81.8 & 88.0 \\
        MSR-VTT & V2T & 49.9 & 57.1 & +7.2 & 81.6 & 88.7 \\
        \addlinespace[1pt]
        MSR-VTT (FT) & T2V & 56.2 & 59.8 & +3.6 & -- & -- \\
        MSR-VTT (FT) & V2T & 55.1 & 58.8 & +3.7 & -- & -- \\
        \addlinespace[1pt]
        MSVD & T2V & 51.9 & 56.8 & +4.9 & 83.0 & 90.1 \\
        MSVD & V2T & 76.7 & 76.9 & +0.2 & 93.8 & 96.4 \\
        \addlinespace[1pt]
        DiDeMo & T2V & 47.7 & 56.6 & +8.9 & 79.8 & 86.8 \\
        DiDeMo & V2T & 47.9 & 54.8 & +6.9 & 80.8 & 86.4 \\
        \addlinespace[1pt]
        ActivityNet & T2V & 46.4 & 51.6 & +5.2 & 79.9 & 89.3 \\
        ActivityNet & V2T & 48.1 & 49.9 & +1.8 & 79.1 & 89.5 \\
        \bottomrule
    \end{tabularx}
    \end{editblock}
    \arrayrulecolor{black}
    \endgroup
\end{table}

\FloatBarrier
\subsection{\edit{Sensitivity to the Candidate-Pool Size \texorpdfstring{$N_{\mathrm{c}}$}{N-c}}}
\label{app:k_sensitivity}

\begin{figure}[htbp]
    \centering
    \includegraphics[width=\linewidth]{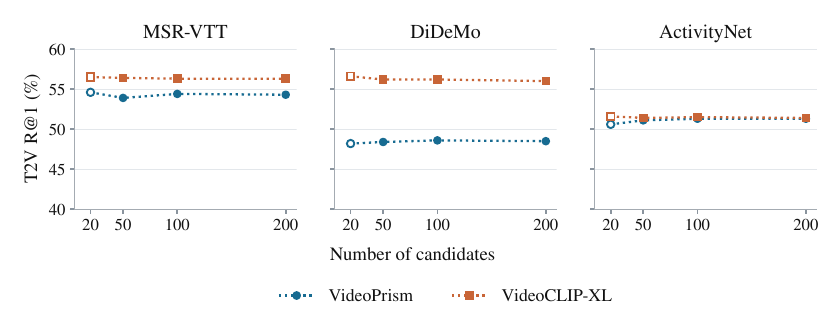}
    \caption{\edit{T2V R@1 (\%) versus candidate-pool size. Each panel uses the same linear recall axis starting at 40\%. Hollow markers at 20 candidates repeat main results; filled markers at other candidate counts are single-run evaluations. Dotted lines guide the eye.}}
    \label{fig:app_k_sensitivity_t2v}
\end{figure}

\edit{The candidate-pool sweep measures how retrieval changes as the reranker processes more candidates.} \edit{Figure~\ref{fig:app_k_sensitivity_t2v} reports T2V R@1 on three datasets with two first-stage retrievers, using 20 to 200 candidates.} \edit{Entries at 20 candidates repeat the main results; the remaining entries are single-run evaluations.}

\edit{Adding candidates does not uniformly improve R@1. MSR-VTT/VP declines at 50 candidates and then partly recovers, whereas ActivityNet/VP improves through 100 candidates.}

\FloatBarrier
\subsection{16-Token Precision Details}
\label{app:token_precision_details}

\edit{Table~\ref{tab:app_token_precision_details} extends the cache-size study in Section~\ref{subsec:token_budget} with a precision comparison at 16 tokens, with all recall values in percent.}

\begin{table}[htbp]
    \centering
    \caption{\edit{MSR-VTT 1ka Recall@K for 16 cached tokens under different cache precisions, in percent. All rows use one cached token per sampled frame.}}
    \label{tab:app_token_precision_details}
    \begingroup
    \normalsize
    \setlength{\tabcolsep}{5pt}
    \renewcommand{\arraystretch}{1.12}
    \arrayrulecolor{gray!70}

\begin{editblock}
\begin{tabularx}{\linewidth}{>{\raggedright\arraybackslash}X c c c c c c}
        \toprule
        \textbf{Precision} & \textbf{\shortstack{T2V\\R@1}} & \textbf{\shortstack{T2V\\R@5}} & \textbf{\shortstack{T2V\\R@10}} & \textbf{\shortstack{V2T\\R@1}} & \textbf{\shortstack{V2T\\R@5}} & \textbf{\shortstack{V2T\\R@10}} \\
        \midrule
        \oursrow
        \textbf{BF16 reference} & \textbf{54.4} & \textbf{78.9} & \textbf{85.6} & 53.1 & \textbf{78.4} & \textbf{86.5} \\
        \oursrow
        FP8-E4M3 & \textbf{54.4} & \textbf{78.9} & 85.5 & \textbf{53.2} & \textbf{78.4} & \textbf{86.5} \\
        \oursrow
        FP4-E2M1 & 54.0 & 78.8 & 85.5 & \textbf{53.2} & \textbf{78.4} & \textbf{86.5} \\
        \bottomrule
    \end{tabularx}
\end{editblock}

    \arrayrulecolor{black}
    \endgroup
\end{table}

The higher-rank cutoffs reinforce the R@1 reading. FP8 leaves T2V and V2T values effectively unchanged from the BF16 reference, and FP4 keeps R@5 and R@10 within 0.1 percentage point on T2V while matching or slightly improving the V2T R@1 reference. \edit{At this cache budget, the tested precision reductions have a small effect on recall.}

\FloatBarrier
\subsection{16-Token Future-Delta Details}
\label{app:future_delta_16token_msr_vtt}

Table~\ref{tab:app_future_delta_16token_msr_vtt} gives Recall@K for the MSR-VTT 16-token future-delta comparison from Section~\ref{subsec:future_delta_compression}, in percent; gains are percentage points. At one cached token per sampled frame, \deltaloss{} improves every reported cutoff in both retrieval directions, with the largest gain at T2V R@1.

\begin{table}[htbp]
    \centering
    \caption{\edit{MSR-VTT 1ka Recall@K for 16 cached tokens with and without \deltaloss{}, in percent. All rows use one cached token per sampled frame.}}
    \label{tab:app_future_delta_16token_msr_vtt}
    \begingroup
    \normalsize
    \setlength{\tabcolsep}{4pt}
    \renewcommand{\arraystretch}{1.12}
    \arrayrulecolor{gray!70}

\begin{editblock}
\begin{tabularx}{\linewidth}{>{\raggedright\arraybackslash}X c c c c c c}
        \toprule
        \textbf{Variant} & \textbf{\shortstack{T2V\\R@1}} & \textbf{\shortstack{T2V\\R@5}} & \textbf{\shortstack{T2V\\R@10}} & \textbf{\shortstack{V2T\\R@1}} & \textbf{\shortstack{V2T\\R@5}} & \textbf{\shortstack{V2T\\R@10}} \\
        \midrule
        16 tokens w/o \deltaloss{} & 52.5 & 78.1 & 84.5 & 52.6 & 77.8 & 85.7 \\
        \oursrow
        \textbf{16 tokens with \deltaloss{}} & \textbf{54.4} & \textbf{78.9} & \textbf{85.6} & \textbf{53.1} & \textbf{78.4} & \textbf{86.5} \\
        \bottomrule
    \end{tabularx}
\end{editblock}

    \arrayrulecolor{black}
    \endgroup
\end{table}

The 16-token results extend the R@1 claim in the main paper to higher cutoffs. The top-rank T2V gain is $+1.9$ percentage points; improvements persist at $+0.8$ R@5 and $+1.1$ R@10. V2T gains are smaller but consistently positive, from $+0.5$ R@1 to $+0.8$ R@10.

\FloatBarrier
\subsection{\edit{Auxiliary Prediction Targets}}
\label{app:future_delta_absolute}
\label{app:auxiliary_targets}

The default \deltaloss{} regresses the patch-level \emph{difference} $\Delta_{t,h}{=}X_{t+h}{-}X_t$. \edit{Table~\ref{tab:auxiliary_targets} compares this target with reconstructing the current-frame feature $X_t$ and predicting the absolute future feature $X_{t+h}$, using 64 cached tokens. The absolute-future-feature control holds the predictor architecture, horizon set, loss weight, and training schedule fixed.}

\begin{table}[htbp]
    \centering
    \caption{\edit{Auxiliary prediction targets with 64 cached tokens on MSR-VTT 1ka. Recall values are percentages. All reconstruction recalls and delta R@1 are mean recall values.}}
    \label{tab:app_future_delta_absolute}
\label{tab:auxiliary_targets}
    \begingroup
    \normalsize
    \setlength{\tabcolsep}{4pt}
    \renewcommand{\arraystretch}{1.12}
    \arrayrulecolor{gray!70}

\begin{editblock}
\begin{tabularx}{\linewidth}{@{}l *{2}{>{\centering\arraybackslash}X} >{\columncolor{ourmethodrow}[\tabcolsep][0pt]\centering\arraybackslash}X@{}}
        \toprule
        \textbf{Recall} & \textbf{\shortstack{Current-frame\\reconstruction $X_t$}} & \textbf{\shortstack{Absolute future\\feature $X_{t+h}$}} & \textbf{\shortstack{Future delta $\Delta_{t,h}$\\(default)}} \\
        \midrule
        T2V R@1 & $54.2$ & 52.8 & $\mathbf{54.6}$ \\
        T2V R@5 & $75.7$ & \textbf{78.2} & 76.9 \\
        T2V R@10 & $84.8$ & 85.3 & \textbf{85.4} \\
        \midrule
        V2T R@1 & $53.9$ & 53.8 & $\mathbf{54.6}$ \\
        V2T R@5 & $77.9$ & 78.6 & \textbf{79.3} \\
        V2T R@10 & $85.8$ & 86.5 & \textbf{86.7} \\
        \bottomrule
    \end{tabularx}
\end{editblock}

    \arrayrulecolor{black}
    \endgroup
\end{table}

\edit{Current-frame reconstruction reaches mean R@1 of 54.2 in T2V and 53.9 in V2T. The delta target is higher by 0.4 and 0.7 points, respectively, favoring feature-change prediction over the tested reconstruction target. The delta target also gives higher R@1 than the absolute-future-feature target ($+1.8$ T2V, $+0.8$ V2T), while the absolute-future target is 1.3 points higher at T2V R@5.}

\FloatBarrier
\Needspace{7\baselineskip}
\subsection{Training-Loss Stack Ablation}
\label{app:loss_stack_ablation}

\edit{Table~\ref{tab:app_loss_stack_ablation} measures the incremental effect of adding training objectives in sequence.} The stage-1 row reports the unreordered VideoPrism~\citep{zhao2024videoprism} retriever. The remaining rows use the same VideoPrism~\citep{zhao2024videoprism} candidate generator and cached joint-reranking setup while turning the training losses on in stages.

\begin{table}[htbp]
    \centering
    \caption{\edit{VideoPrism\methodcite{zhao2024videoprism} loss-stack ablation on MSR-VTT 1ka, with Recall@$K$ in percent. The stage-1 row reports the VideoPrism retriever before reranking. Reranker rows add the training objectives in stages; the full \architecture{} row adds \deltaloss{} on top of VTM, MLM, and VTC.}}
    \label{tab:app_loss_stack_ablation}
    \begingroup
    \normalsize
    \setlength{\tabcolsep}{3pt}
    \renewcommand{\arraystretch}{1.12}
    \arrayrulecolor{gray!70}

\begin{editblock}
\begin{tabularx}{\linewidth}{>{\raggedright\arraybackslash}X r r r r r r}
        \toprule
        \textbf{Active training objectives} & \textbf{\shortstack{T2V\\R@1}} & \textbf{\shortstack{T2V\\R@5}} & \textbf{\shortstack{T2V\\R@10}} & \textbf{\shortstack{V2T\\R@1}} & \textbf{\shortstack{V2T\\R@5}} & \textbf{\shortstack{V2T\\R@10}} \\
        \midrule
        \methodcell{Stage-1 only}{zhao2024videoprism}\newline{}No reranker losses & 50.1 & -- & -- & 49.8 & -- & -- \\
        $\mathcal{L}_{\mathrm{vtm}}$ & 52.8 & 76.9 & 84.0 & 50.6 & 77.4 & 85.2 \\
        $\mathcal{L}_{\mathrm{vtm}}+\mathcal{L}_{\mathrm{mlm}}$ & 52.7 & 76.2 & 84.2 & 51.4 & 76.0 & 84.9 \\
        $\mathcal{L}_{\mathrm{vtm}}+\mathcal{L}_{\mathrm{mlm}}+\mathcal{L}_{\mathrm{vtc}}$ & 54.0 & \textbf{77.8} & 85.3 & 53.6 & 77.9 & \textbf{87.0} \\
        \oursrow
        Full: \textbf{$\mathcal{L}_{\mathrm{vtm}}+\mathcal{L}_{\mathrm{mlm}}+\mathcal{L}_{\mathrm{vtc}}+\deltaloss$} & \textbf{54.6} & 76.9 & \textbf{85.4} & \textbf{54.6} & \textbf{79.3} & 86.7 \\
        \bottomrule
    \end{tabularx}
\end{editblock}

    \arrayrulecolor{black}
    \endgroup
\end{table}

\edit{VTM supplies an initial gain of $+2.7$ T2V R@1 over the VideoPrism~\citep{zhao2024videoprism} prior.} \edit{Adding MLM changes T2V/V2T R@1 by $-0.1/+0.8$ points, followed by $+1.3/+2.2$ from VTC and $+0.6/+1.0$ from \deltaloss{}.} \edit{Higher-rank recalls are mixed: adding \deltaloss{} lowers T2V R@5 by 0.9 points.} \edit{Each increment is conditional on the losses already included.}

\FloatBarrier
\subsection{Future-Delta Loss Horizon Sweep}
\label{app:horizon_sweep}

\edit{We selected $h{=}3$ by three-fold cross-validation on the MSR-VTT training set, using held-out training folds for validation and leaving the evaluation set untouched. This horizon gave the most consistent validation R@1 across folds, with stability assessed by its standard deviation. The test-set sweep in Table~\ref{tab:app_horizon_sweep} shows sensitivity to the prediction offset: among the tested horizons, $h{=}3$ improves R@1 over the no-auxiliary reference in both retrieval directions.}

\begin{table}[htbp]
    \centering
    \caption{\edit{Future-delta horizon ablation on MSR-VTT 1ka. Recall values are percentages; the no-auxiliary reference in Table~\ref{tab:app_loss_stack_ablation} has T2V/V2T R@1 of 54.0/53.6.}}
    \label{tab:app_horizon_sweep}
    \begingroup
    \normalsize
    \setlength{\tabcolsep}{5pt}
    \renewcommand{\arraystretch}{1.15}
    \arrayrulecolor{gray!70}

\begin{editblock}
\begin{tabularx}{\linewidth}{>{\raggedright\arraybackslash}X c c c c c c}
        \toprule
        \textbf{Horizon} & \textbf{\shortstack{T2V\\R@1}} & \textbf{\shortstack{T2V\\R@5}} & \textbf{\shortstack{T2V\\R@10}} & \textbf{\shortstack{V2T\\R@1}} & \textbf{\shortstack{V2T\\R@5}} & \textbf{\shortstack{V2T\\R@10}} \\
        \midrule
        $h=2$ & 53.8 & 78.2 & 85.5 & 53.0 & 78.7 & 86.6 \\
        \oursrow
        \textbf{$h=3$ (default)} & \textbf{54.6} & 76.9 & 85.4 & \textbf{54.6} & \textbf{79.3} & 86.7 \\
        $h=4$ & 54.4 & \textbf{78.7} & \textbf{85.6} & 53.0 & 78.2 & \textbf{86.9} \\
        \bottomrule
    \end{tabularx}
\end{editblock}

    \arrayrulecolor{black}
    \endgroup
\end{table}

\edit{No horizon dominates every R@K cutoff.} \edit{Relative to the no-auxiliary reference of 54.0/53.6 T2V/V2T R@1, $h{=}2$ gives 53.8/53.0 and $h{=}4$ gives 54.4/53.0.} \edit{The benefit therefore depends on the offset; the default $h{=}3$ gives the strongest R@1 result in both directions.}

\FloatBarrier
\Needspace{13\baselineskip}
\subsection{Reranker Text-Encoder Control}
\label{app:secondary_controls}

\edit{This control measures the effect of increasing reranker capacity while retaining the cache and training setup.} \edit{Replacing the default 33M joint encoder with BERT-base~\citep{devlin2019bert} (110M parameters) adds 77M reranker parameters, a $3.3\times$ increase, but yields small mixed changes on MSR-VTT 1ka.}

\begin{table}[htbp]
    \centering
    \caption{\edit{MSR-VTT 1ka reranker-backbone control, with Recall@$K$ in percent. The BERT-base\methodcite{devlin2019bert} control raises the added reranker size from 33M to 110M parameters.}}
    \label{tab:app_secondary_controls}
    \begingroup
    \normalsize
    \setlength{\tabcolsep}{3pt}
    \renewcommand{\arraystretch}{1.15}
    \arrayrulecolor{gray!70}

\begin{editblock}
\begin{tabularx}{\linewidth}{>{\raggedright\arraybackslash}X c c c c c c c}
        \toprule
        \textbf{Variant} & \textbf{\shortstack{Added reranker\\params}} & \textbf{\shortstack{T2V\\R@1}} & \textbf{\shortstack{T2V\\R@5}} & \textbf{\shortstack{T2V\\R@10}} & \textbf{\shortstack{V2T\\R@1}} & \textbf{\shortstack{V2T\\R@5}} & \textbf{\shortstack{V2T\\R@10}} \\
        \midrule
        \oursrow
        \textbf{Default \architecture{}} & \textbf{33M} & \textbf{54.6} & 76.9 & 85.4 & \textbf{54.6} & \textbf{79.3} & \textbf{86.7} \\
        \methodcell{BERT-base text encoder}{devlin2019bert} & 110M & 52.9 & \textbf{77.5} & \textbf{85.5} & 54.0 & 77.9 & 86.5 \\
        \bottomrule
    \end{tabularx}
\end{editblock}

    \arrayrulecolor{black}
    \endgroup
\end{table}

\edit{BERT-base~\citep{devlin2019bert} improves T2V R@5 and R@10 by 0.6 and 0.1 percentage points but lowers R@1 by 1.7 (T2V) and 0.6 (V2T) percentage points despite using $3.3\times$ more reranker parameters.} \edit{Increasing capacity under this training schedule does not improve R@1.}

\FloatBarrier
\section{\edit{Comparison with published systems}}
\label{app:comparison_context}

\edit{\paragraph{Accuracy and model footprint.} Table~\ref{tab:competitor_accuracy} and Figure~\ref{fig:method_positioning} compare MSR-VTT T2V R@1 with the parameter footprint of query-time language and scoring modules.} \edit{Each system retains its published training and evaluation protocol.} \edit{The main-text controls separately measure the effects of cache construction and scoring on shared candidates or cached tokens.}

\begin{table}[htbp]
\centering
\caption{\edit{MSR-VTT T2V R@1 and online parameters. Panels identify the counted modules. Public-architecture counts are rounded to 0.1M; VEDJE totals retain the reported whole-million estimates. Video-ColBERT counts the full SigLIP text module. MLLM rows count the base decoder; their pipelines also use first-stage query encoding and fine-tuning components. A dash denotes an unestablished count or inapplicable entry. ZS/FT denote zero-shot/fine-tuned stage 1; parentheses give matched gains.}}
\label{tab:competitor_accuracy}

\begin{editblock}
\begin{tabularx}{\linewidth}{X rr}
\toprule
Method & \shortstack{Online\\parameters} & R@1 (\%) \\
\midrule
\multicolumn{3}{l}{\textbf{Text encoding and learned scoring}} \\
CLIP4Clip B/32 & 63.4M & 44.5 \\
CLIP4Clip B/16 (reproduced) & 63.4M & 46.4 \\
X-CLIP B/16 & 64.0M & 49.3 \\
EERCF B/16 & 63.7M & 49.9 \\
Video-ColBERT SigLIP-B/16 & 110.3M & 51.5 \\
CrossTVR-Large & -- & 54.0 \\
\midrule
\multicolumn{3}{l}{\textbf{First-stage text encoders}} \\
VideoPrism stage 1 & 109.6M & 50.1 \\
VideoCLIP-XL stage 1 (ZS) & 124.0M & 50.1 \\
VideoCLIP-XL stage 1 (FT) & 124.0M & 56.2 \\
PE-Core-B stage 1 & 354.0M & 47.6 \\
\midrule
\multicolumn{3}{l}{\textbf{VEDJE: estimated online totals}} \\
\oursrow VEDJE$_\mathrm{VP}$ & 143M & $54.6\pm0.1$ (+4.5) \\
\oursrow VEDJE$_\mathrm{VCLIP}$ (ZS) & 157M & $56.5\pm0.2$ (+6.4) \\
\oursrow VEDJE$_\mathrm{VCLIP}$ (FT) & 157M & $59.8\pm0.2$ (+3.6) \\
\oursrow VEDJE$_\mathrm{PE}$ & 387M & $53.9\pm0.2$ (+6.3) \\
\midrule
\multicolumn{3}{l}{\textbf{Multimodal rerankers: base language decoder}} \\
LamRA (reproduced) & 7.6B & 59.7 \\
CaRe-DPO & 7.6B & 64.1 \\
\bottomrule
\end{tabularx}
\end{editblock}

\end{table}

\edit{\paragraph{Query-time parameter accounting.} For indexed retrieval, CLIP4Clip~\citep{luo2022clip4clip}, X-CLIP~\citep{ma2022x}, EERCF~\citep{tian2024towards}, and Video-ColBERT~\citep{Reddy_2025_CVPR} can reuse video representations computed independently of the text.} \edit{We count text-encoding and learned-scoring weights from their public architectures.} \edit{Figure~\ref{fig:method_positioning} uses the reproduced CLIP4Clip B/16 result of 46.4 from \citet{tian2024towards}, labeled as a reproduction in its source table.} \edit{The CLIP text tower and logit scale contain approximately 63.4M parameters. X-CLIP's scoring matrices add approximately 0.5M parameters; EERCF's coarse-score and frame-score matrices add approximately 0.3M, alongside its parameter-free text-gated interaction block.} \edit{The Video-ColBERT count covers the full 110.3M-parameter SigLIP-B/16 text module, including its output head; the paper does not specify whether its tokenwise scorer uses that head.}

\edit{\paragraph{VEDJE parameter totals.} The total combines the first-stage text encoder with the approximately 33M joint reranker and its small scoring heads.} \edit{Public text-encoder architectures contain approximately 109.6M parameters for VideoPrism~\citep{zhao2024videoprism}, 124.0M for VideoCLIP-XL~\citep{wang_etal_2024_videoclip}, and 354.0M for PE-Core-B/16~\citep{bolya2025perceptionencoder}.} \edit{Table~\ref{tab:competitor_accuracy} retains the reported VEDJE total estimates of 143M, 157M, and 387M.} \edit{The visual backbone, compressor, and training-only predictor run during indexing or training.}

\edit{\paragraph{Multimodal rerankers.} The LamRA reproduction and CaRe-DPO results are reported together by \citet{lee2025captioning}, using VideoChat-Flash-7B.} \edit{The public model's Qwen2 base decoder contains approximately 7.6B parameters, including its untied input embeddings and output head. This is the component counted in Table~\ref{tab:competitor_accuracy} and Figure~\ref{fig:method_positioning}.} \edit{The complete pipelines also use a first-stage query encoder and fine-tuning components, including any unmerged adapters and CaRe-DPO's role embeddings.} \edit{The reproduced LamRA result is 59.7 R@1, compared with 59.8 for VEDJE at approximately 157M online parameters; the base decoder alone has about 49 times that parameter count.} \edit{CaRe-DPO reaches higher recall at 64.1 and remains visible in the comparison.} \edit{CrossTVR-Large's~\citep{DAI2025111099} online/offline parameter partition was not established, so its accuracy is listed without a cost coordinate.}

\edit{\paragraph{Interpreting model footprint.} Parameter footprint describes the size of the models used during a query.} \edit{Operation count, sequence length, candidate count, hardware, and cache access also determine execution time.} \edit{Section~\ref{subsec:query_cost} and Appendix~\ref{app:latency_io_protocol} report measured VEDJE query latency, including text encoding, search, cache access, and scoring.} \edit{Figure~\ref{fig:method_positioning} provides the broader comparison under the stated parameter accounting.}

\end{document}